\pdfoutput=1

\documentclass[11pt]{article}

\usepackage[final]{coling}

\usepackage{times}
\usepackage{latexsym}

\usepackage[T1]{fontenc}

\usepackage[utf8]{inputenc}

\usepackage{microtype}

\usepackage{inconsolata}

\usepackage{graphicx}

\usepackage{multirow}
\usepackage{array}
\usepackage{amsmath}
\usepackage{xcolor}
\usepackage{colortbl}
\usepackage{bbding}
\usepackage{float}
\usepackage{makecell}
\usepackage{needspace}
\usepackage{color,soul}
\usepackage{subcaption}

\graphicspath{ {./images/} }

\definecolor{cons_c}{HTML}{7EA1CC}
\definecolor{stance_c}{HTML}{7ECC8E}
\definecolor{target_c}{HTML}{A07ECC}
\definecolor{reason_c}{HTML}{CCB77E}

\newcolumntype{?}{!{\vrule width 1pt}}
\newcolumntype{P}[1]{>{\centering\arraybackslash}m{#1}}

\definecolor{babyblue}{rgb}{0.63, 0.79, 0.95}
\definecolor{brickred}{rgb}{0.9, 0.35, 0.43}

\title{Summarization of Opinionated Political Documents with Varied Perspectives}

\author{
 \textbf{Nicholas Deas}, %
 \textbf{Kathleen McKeown}%
\\
\\
 Department of Computer Science, Columbia University
\\
 \small{
   \href{mailto:ndeas@cs.columbia.edu}{[ndeas,kathy]@cs.columbia.edu}
 }
}

\begin{document}
\maketitle
\begin{abstract}
    Global partisan hostility and polarization has increased, and this polarization is heightened around presidential elections.
    Models capable of generating accurate summaries of diverse perspectives can help reduce such polarization by exposing users to alternative perspectives.
    In this work, we introduce a novel dataset and task for independently summarizing each political perspective in a set of passages from opinionated news articles. 
    For this task, we propose a framework for evaluating different dimensions of perspective summary performance.
    We benchmark 11 summarization models and LLMs of varying sizes and architectures through both automatic and human evaluation. While recent models like GPT-4o perform well on this task, we find that all models struggle to generate summaries that are faithful to the intended perspective. 
    Our analysis of summaries focuses on how extraction behavior is impacted by features of the input documents\footnote{We make our code available at \url{https://github.com/NickDeas/PoliSum}}.
\end{abstract}

\section{Introduction}

    Political ideologies can lead people to develop misperceptions of groups with opposing opinions \cite{chambers-misperceptions}, and political events, such as the 2024 US presidential election, French legislative election, or the Brexit referendum, can reinforce negative attitudes \cite{hanna-election,wellings-brexit}. 
    These misperceptions, reinforced by news consumption on social media \cite{levy-media}, contribute to increased polarization and instability \cite{braley-democracy,lees-polarization}. 
    Exposure to alternative perspectives, however, has been shown to help alleviate polarization \cite{balietti-exposure}. To encourage such exposure, some groups focus on aggregating (e.g., All Sides\footnote{\url{https://www.allsides.com/}}) or summarizing different political perspectives on divisive issues (e.g., The Flip Side\footnote{\url{https://www.The Flip Side.io/}}, Ground News\footnote{\url{https://ground.news/}}).

    Large language models' (LLMs) ability to summarize opinions and news has recently neared human performance \cite{zhang-benchmarking,bhaskar-prompted}. 
    Recent work, however, has also shown that LLMs can unfairly represent diverse opinions in review and tweet summarization \cite{zhang-fair, huang-bias, tay-review}.

    \begin{figure}
        \centering
        \includegraphics[width=0.48\textwidth]{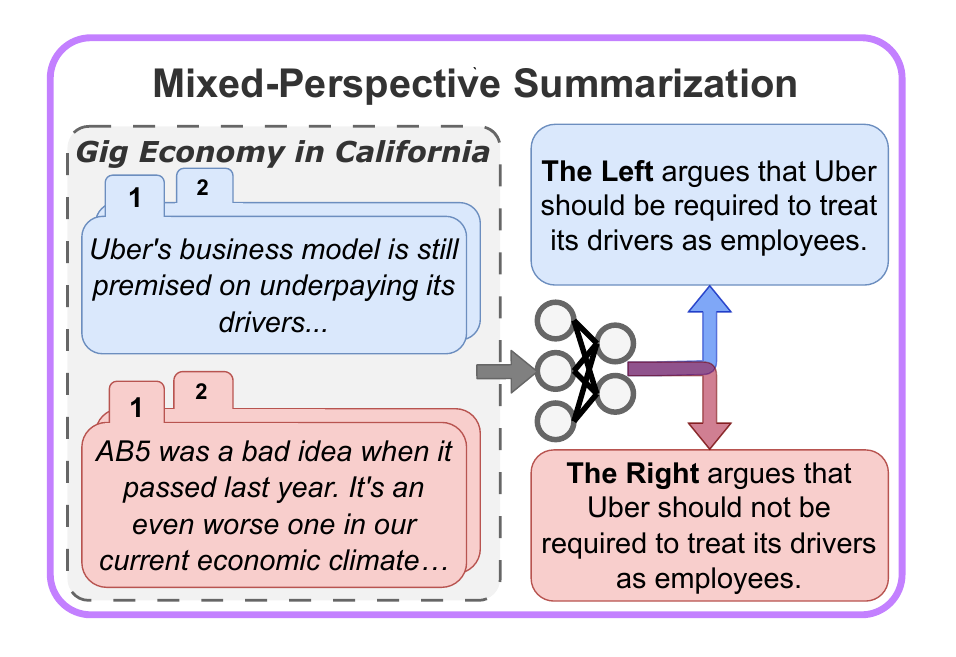}
        \caption{Example of the Mixed-Perspective Setting of \textsc{PoliSum}. Given a set of opinionated passages reflecting a mix of views, \textsc{PoliSum} requires models to generate both a summary of the left and right-leaning political perspectives. 
        } 
        \label{fig:task-diagram}
    \end{figure}

    To evaluate current models' capabilities in summarizing multiple perspectives in a set of input documents, we propose a novel dataset, \textsc{PoliSum}, and task, mixed perspective summarization\footnote{We make our dataset available to researchers who have been granted permission to use the data for their work by The Flip Side.}. 
    Our task involves generating independent abstractive summaries of both left and right-leaning political perspectives given collections of opinionated news passages with mixed perspectives. Mixed perspective summarization thus requires models to both distinguish the perspectives reflected in input texts as well as produce a pair of summaries representing those differing perspectives. 
    We summarize our primary contributions as follows: 
    \begin{enumerate}
        \vspace{-.4em}
        \itemsep-.3em
            \item We introduce a novel benchmark dataset, \textsc{PoliSum}, for mixed perspective summarization, requiring models to independently summarize political perspectives on controversial issues, provided passages from news and op-eds.
            \item We introduce an initial framework for evaluating different components of perspective summaries, including both human evaluation and automatic metrics. 
            \item We benchmark 11 models, finding that most approaches struggle to faithfully summarize the intended perspective and that improving automatic evaluation metrics is an important area for future work.
            \item We analyze model extraction behavior (e.g., which documents models extract from and to what degree) and find that generated summaries suffer from biases due to input document position, document length, and use of arousing terms.
    \end{enumerate}

\section{\textsc{PoliSum} Dataset}

    \begin{table*}[!htbp]
        \centering
        \small
        \begin{tabular}{| c | P{1.5cm} P{2cm} P{2cm} | P{1cm} | P{1cm} | c c c |}
             \hline
             \multirow{2}{*}{Dataset} & \multirow{2}{*}{Domain} & \multirow{2}{*}{Multi-document} & \multirow{2}{*}{Multi-summary} & \multirow{2}{1cm}{\centering \# Gold} & \multirow{2}{1cm}{\centering Gold Length} & \multicolumn{3}{c|}{\% Novel n-grams} \\
             & & & & & & 1 & 2 & 3 \\
             \hline
             XSUM                        & News & \XSolidBrush & \XSolidBrush & 11,334 &  23.3 & 35.8 & 83.5 & 95.5 \\
             Multi-News                  & News & \Checkmark & \XSolidBrush & 5,622  & 263.7 & 17.8 & 57.1 & 75.7 \\
             \textsc{CoCoTrip}           & Reviews & \Checkmark & \Checkmark & $48 \cdot 3$ & 132.9 & 22.8 & 72.4 & 91.4 \\
             \hline
             \textsc{PoliSum}            & Op-Eds & \Checkmark& \Checkmark & $907 \cdot 2$   & 19.7  & 40.4 & 86.8 & 97.5 \\
             \hline
        \end{tabular}
        \caption{Summary statistics for \textsc{PoliSum} and existing summarization datasets, including data size, reference length, and percentage of novel n-grams in the summary compared to source documents. \# Gold for multi-summary datasets is presented as \# \textit{Samples} $\cdot$ \# \textit{References/Sample}. CoCoTrip statistics referenced from \citet{comparative-opinion-summ}.}
        \label{tab:data-stats}
    \end{table*}

    \textbf{Data Source.} To enable evaluating political perspective summaries, we collect opinionated texts and paired summaries from The Flip Side\footnote{We thank The Flip Side for generously allowing us to use their archive for this work: \url{https://www.The Flip Side.io/archives}}. Since September 2017, The Flip Side has covered politically controversial issues (i.e. overturning of Roe v. Wade, presidential elections) in the United States. Toward bridging political divides, each publication includes a headline reflecting the central issue, curated passages authored by political commentators with varying political leaning, and a pair of short summaries of the left and right perspectives\footnote{Some publications lack individual perspective summaries, primarily on topics where both parties largely agree or for meta-articles that do not cover political topics. These samples are not included in the dataset.}. We collect this information from The Flip Side archive, gathering nearly 1,000 samples. An excerpted example of source passages and summaries is shown in \autoref{tab:data-ex}.

    \begin{table}[htbp]
        \centering
        \scriptsize
        \begin{tabular}{| P{1.2cm} | P{2.55cm} | P{2.55cm} |}
            \hline
             Headline & \multicolumn{2}{c|}{\textbf{Hollywood Strike}} \\
             \hline
             Perspective & Left & Right \\
             \hline
             Source Passages & ...This is a huge issue, but there's another one — AI... A lot of the TV episodes and movies produced by Hollywood are, by nature, highly formulaic... & ...By replacing the once majority straight white cast with a ridiculously disproportionate percentage of minorities, industry players alienated a commensurate portion of the audience...\\
             \hline
             Summaries & The left supports the strikers, arguing that AI will soon be a threat to workers in many industries. & The right argues that Hollywood’s troubles stem from its embrace of leftist politics.\\
             \hline
        \end{tabular}
        \caption{Example of partial input passages and paired perspective summaries in \textsc{PoliSum}.}
        \label{tab:data-ex}
    \end{table}

    Table 1 presents summary statistics of the collected multi-document, multi-summary dataset (data statement included in \autoref{app:data-statement}).  We also show statistics for \textit{\textsc{CoCoTrip}} (\citealt{comparative-opinion-summ}; the most comparable multi-document, multi-summary dataset), \textit{XSUM} (\citealt{narayan-xsum}; a single-document, single-summary corpus with similarly short references), and \textit{Multi-News} (\citealt{multinews}, a multi-document, single-summary news corpus). The aim of \textsc{CoCoTrip} is to generate multiple hotel review summaries for a pair of comparable hotels: two summaries highlighting each hotel's unique attributes, and one highlighting similarities. In contrast, \textsc{PoliSum} contains almost 20 times more labeled examples and focuses on political perspectives rather than reviews. Additionally, compared to XSum, \textsc{PoliSum} is similar in length but slightly more abstractive, measured by n-gram overlap between source and reference. 
    On average, there are 9.6 source passages for each publication (4.8 per perspective), with a combined length comparable to a news article ($\sim$1,300 tokens). 

    \textbf{Perspective Summarization Task.} Using this dataset, the aim of the our proposed task is to generate a pair of summaries (i.e., \textit{Summaries} in \autoref{tab:data-ex}) representing the left and right political perspectives on a controversial topic (i.e. \textit{Headline}). As input, models are provided with two sets of opinionated editorial passages (i.e., \textit{Source Passages}), each representing one of the two political perspectives. Thus, the dataset enables a multi-document, multi-summary task. 
    As with traditional news summarization tasks, the summary for each perspective should condense the information present in the passages. At the same time, the generated left and right-perspective summaries should faithfully represent the perspectives reflected in their respective passages.

\section{Experiments}

    Traditional opinion summarization typically involves summarizing a common perspective among all inputs, 
    which we refer to as the Single-Perspective setting\footnote{Experimental details and results of the Single-Perspective task on \textsc{PoliSum} are included in \autoref{app:single-persp}}.
    We instead introduce an additional challenge requiring models to identify input documents relevant to the intended perspective,
    referred to as the Mixed-Perspective setting and illustrated in \autoref{fig:task-diagram}. In this setting, approaches are provided with a mixed set of left and right-leaning passages as input and are tasked with producing independent left and right-perspective summaries from the same input. On average, sets of inputs contain $\sim$10 passages total, $\sim$5 passages for each perspective.
        
    Model inputs are constructed by concatenating all passages, ensuring that the left and right perspectives alternate when possible. 
    Due to position biases in summarization \cite{chhabra-revisiting}, the perspective reflected in the first input document may spuriously impact the resulting summary quality. To mitigate these effects, we report average performance between two variations of the input data. First, models are evaluated on inputs beginning with a left-perspective passage followed by alternating perspectives (e.g., left, right, left, etc.), and then on inputs beginning with a right-perspective passage followed by alternating perspectives (e.g., right, left, right, etc.). The order of individual passages that share a perspective is randomized.
    Because the reference summaries are limited to a single sentence, all model generations are truncated to the first sentence before evaluation.

    \subsection{Models} 
        A variety of summarization models and instruction-tuned LLMs are considered. 

        \textbf{Extractive Upper Bound.} 
            To represent potential extractive summarization approaches, we estimate an upper bound performance based on semantic similarity to the reference summaries. We consider all sentences contained in the source passages to be potential extractive summaries. We then select the sentence with the highest \textsc{BERTScore} relative to the reference summary and calculate the remaining coverage and faithfulness metrics. 

        \textbf{Neural Summarization Baselines.} 
            To represent neural summarization models, we evaluate the large variants of BART \cite{bart}, T5 \cite{t5}, and Flan-T5 \cite{chung-flan}. As BART is not pretrained on any summarization task, we use BART finetuned on the CNN/DailyMail (CNN/DM; \citealt{hermann-cnndm, nallapati-cnndm}) dataset. In contrast to \textsc{PoliSum}, CNN/DM and other prior summarization datasets lack summaries for individual perspectives. So, in order to encourage BART to generate distinct summaries matching the references' format, the BART decoder is forced to begin generations with "The left" or "The right" respectively. T5 and Flan-T5 are prompted to generate summaries of individual perspectives. See \autoref{app:zs-baselines} for other generation hyperparameter details.
            
        \textbf{LLMs.}
            We evaluate 8 additional LLMs of varying sizes and architectures to establish baselines for current state-of-the-art models\footnote{We use instruction-tuned variants of all applicable LLMs, but for simplicity, omit "Instruct" in names throughout results.}.
            \textbf{Mistral-7B} (\textit{mistralai/Mistral-7B-Instruct-v0.1}; \citealt{jiang-mistral}), \textbf{Llama-3.0-8B and 70B} (\textit{meta-llama/Meta-Llama-3-8B-Instruct} and \textit{meta-llama/Meta-Llama-3-70B-Instruct}; \citealt{dubey-llama}), \textbf{Llama-3.3-70B} (\textit{meta-llama/Meta-Llama-3.3-70B-Instruct}), \textbf{Vicuna-7B and 13B} (\textit{lmsys/vicuna-7b-v1.5} and \textit{lmsys/vicuna-13b-v1.5}; \citealt{chiang-vicuna}), \textbf{Mixtral-8x22B} (\textit{mistralai/Mixtral-8x22B-Instruct-v0.1}; \citealt{jiang-mixtral}), and \textbf{GPT-4o} (\textit{gpt-4o-2024-05-13})\footnote{\url{https://openai.com/index/hello-gpt-4o/}}.
            Due to the length of input documents, we evaluate all LLMs in a zero-shot setting. See \autoref{app:prompt-baselines} for additional generation hyperparameters, prompts, and reproducibility details.

    \subsection{Metrics}

        \paragraph{Coverage. }
            Summary coverage is first measured with \textsc{Rouge}-1 and \textsc{Rouge}-L between the generated perspective summaries and their corresponding references \cite{rouge-score}. To account for the weaknesses of \textsc{Rouge}, we also employ \textsc{BERTScore} \cite{zhang-bertscore} and \textsc{Bleurt} \cite{sellam-bleurt} using the \textit{deberta-large-xnli} and \textit{\textsc{Bleurt}-20-D6} checkpoints respectively due to higher correlations with human judgments compared to alternatives\footnote{See human correlations for \textsc{BERTScore} linked in \url{https://github.com/Tiiiger/bert_score} and for \textsc{Bleurt} in \url{https://github.com/google-research/Bleurt/blob/master/checkpoints.md}}. For each metric, we also use a two-sample t-test to determine whether left and right-perspective coverage scores significantly differ ($p\leq0.05$).
        
        \paragraph{Faithfulness. }
            In addition to coverage, we also measure how well the generated summaries are consistent and faithful to the input passages representing the intended perspective. 
            Because faithfulness metrics for our specific task do not exist, we rely on existing factuality metrics as approximations:
            S\textsc{umma}C \cite{laban-summac} and \textsc{AlignScore} \cite{zha-alignscore}.
            S\textsc{umma}C is a NLI-based metric that measures consistency between summaries and source documents. Alternatively, \textsc{AlignScore} uses information alignment 
            to measure factual consistency. 
            All metrics produce a score between 0 and 100 where higher values reflect more consistent and faithful summaries. 
            
        \paragraph{Perspective. }

            \begin{figure}[!htbp]
                \centering
                \vspace{-1.7em}
                \includegraphics[width=0.45\textwidth]{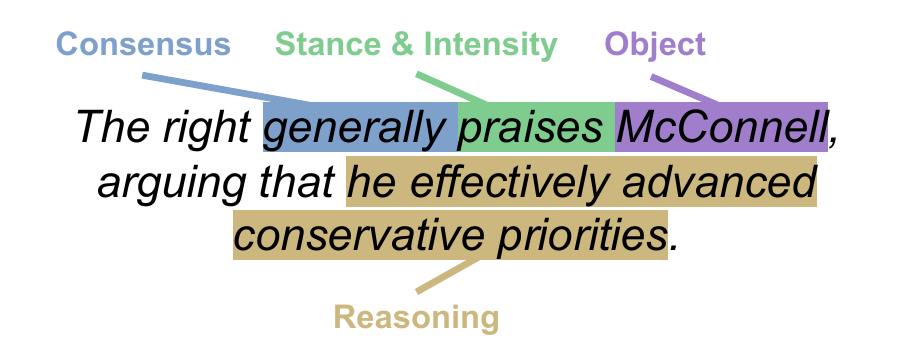}
                \vspace{-1.5em}
                \caption{Example perspective summary with indicators of different perspective dimensions highlighted: {\color{cons_c} \textit{Consensus}}, {\color{stance_c} \textit{Stance and Intensity}}, {\color{target_c} \textit{Object}}, and {\color{reason_c} \textit{Reasoning}}} 
                \label{fig:persp-dims}
            \end{figure}
        
            To measure how well the generated summaries reflect the intended perspective, we decompose dimensions of how perspectives are captured in summaries. \textit{Stance} represents whether the summary accurately portrays the intended perspective's overall attitude toward the object (e.g., for or against), while \textit{Object} represents whether the summary identifies the salient stance object among input documents. These dimensions correspond to the attitude and stance object of the stance triangle theory respectively \cite{dubois-stance-triangle}. Because our task requires expressing perspectives in natural language as opposed to prior stance detection work, we include two additional dimensions: \textit{Intensity} reflects whether the extent to which a perspective is held or the strength of the belief is accurately captured by a perspective summary (e.g., "praises" rather than "supports" may reflect a stronger belief); and \textit{Reasoning} refers to whether the elaboration in a perspective summary is supported by the input documents. Finally, our task requires summarizing multiple perspectives toward a single object, which may not fully agree. Therefore, we include a final dimension to measure whether perspective summaries appropriately express the extent to which input documents agree or disagree, called \textit{Consensus}.

            We conduct a human evaluation of the reference texts and summaries generated by Mistral, Mixtral, and GPT-4o
            to capture these dimensions. We recruit 6 political science students to rate how well a given summary accurately captures each dimension as reflected in the inputs. In total, each of 40 left and 40 right-leaning summaries randomly sampled from the dataset are judged by 2 annotators. 
            
            Additionally, we propose initial metrics to automate selected dimensions drawing on prior work in stance detection and political summarization. To measure \textit{Stance}, we employ JointCL, a zero-shot stance classifier \cite{liang-jointcl,allaway-zssd}.
            Using the headlines as the object, we calculate the Root Mean Squared Error (RMSE) between predictions on the generated summaries and references. To capture \textit{Object}, we use a T5-based keyphrase generation model \cite{li-target-stance} as a zero-shot stance object classifier, and calculate BERTScore between predictions on generated summaries and input passages.
            Finally, similar to \cite{lee-neus}, we measure \textit{Intensity} by calculating the average arousal score of polarizing terms according to the VAD lexicon \cite{mohammad-vad}. Intensity scores are again calculated as the RMSE between arousal scores for generated summaries and input passages.
            Due to their complexity, we leave further investigation of automated evaluations for \textit{Consensus} and \textit{Reasoning} to future work. See \autoref{app:judgments} and \autoref{app:metric-details} for further details on human and automatic metrics respectively.

    \begin{table*}[!htbp]
        \centering
        \scriptsize
        \begin{tabular}{| P{2cm} | P{.6cm} P{.6cm} P{.9cm} P{.6cm} | P{.8cm} P{.6cm} | | P{.6cm} P{.6cm} P{.9cm} P{.6cm} | P{.8cm} P{.6cm} |}
            \hline
            \multirow{3}{*}{\textbf{Model}} & \multicolumn{6}{c||}{\textbf{Left}} & \multicolumn{6}{c|}{\textbf{Right}} \\
                                    & \multicolumn{4}{c|}{Coverage} & \multicolumn{2}{c||}{Faithfulness} & \multicolumn{4}{c|}{Coverage} & \multicolumn{2}{c|}{Faithfulness} \\
                                    & R1   & RL   & \textsc{Bleurt} & BERT& \textsc{SummaC} & Align & R1   & RL    & \textsc{Bleurt} & BERT & \textsc{SummaC} & Align \\
            \hline
            Extractive              & 13.14 & 10.25 & 22.50 & 55.37 & 88.85 & 97.25  & 13.21 & 10.34 & 22.67 & 55.01 & 89.23 & 97.01 \\
            \hline
            BART$_{CNN/DM}$         & 25.3 & 22.7 & \cellcolor{babyblue} 30.5 & \cellcolor{babyblue} \underline{61.6} & 24.8 & 21.0 & 24.7 & 22.3 & 28.9 & 60.0 & \cellcolor{brickred} 25.7 & \cellcolor{brickred} 28.9 \\
            T5                      & 24.7 & 22.6 & 30.9 & 60.8 & 23.3 & 12.0 & \cellcolor{brickred} 25.7 & \cellcolor{brickred} \textbf{23.6} & 31.2 & \underline{61.0} & \cellcolor{brickred} 24.0 & \cellcolor{brickred} 14.7 \\
            Flan-T5                 & 24.1 & 21.2 & 29.6 & 58.9 & 26.2 & 35.8 & 24.9 & 21.8 & 29.3 & 58.8 & \cellcolor{brickred} \underline{26.9} & 37.8 \\
            Mistral                 & 25.7 & 21.4 & \cellcolor{babyblue} 34.5 & \cellcolor{babyblue} 60.4 & 25.5 & 30.1 & 25.6 & 21.3 & 33.2 & 59.3 & 26.5 & \cellcolor{brickred} 40.5 \\
            Mixtral                 & 25.6 & 21.3 & \cellcolor{babyblue} 37.2 & 60.9 & 24.9 & 36.9 & 25.7 & 21.3 & 36.3 & 60.5 & \cellcolor{brickred} 25.2 & \cellcolor{brickred} 51.7 \\
            Vicuna (7B)             &   24.9 &   21.6 &  \cellcolor{babyblue} 33.3 &  \cellcolor{babyblue} 59.6 &  \textbf{27.9} &  30.7 &   25.3 &   21.8 &  32.7 &  58.2 &  \cellcolor{brickred} \textbf{29.4} &  \cellcolor{brickred} 39.2 \\
            Vicuna (13B)            &  \cellcolor{babyblue} \underline{26.4} & \cellcolor{babyblue} \underline{23.0} &  \cellcolor{babyblue} 37.2 & \cellcolor{babyblue} 61.4 &   \underline{26.4} &  37.2 &  25.3 &  22.0 &  35.8 &  60.3 &   26.3 & \cellcolor{brickred} \textbf{54.0} \\
            Llama-3.0 (8B)  &   \textbf{28.0} &   \textbf{23.9} & \cellcolor{babyblue} \underline{38.1} & \cellcolor{babyblue} 60.4 &  25.8 &  35.2 & \textbf{27.7} &   \underline{23.5} &  \underline{37.3} &  59.1 & \cellcolor{brickred} \underline{26.9} & \cellcolor{brickred} 44.8 \\
            Llama-3.0 (70B)           & 23.1 & 19.0 & \cellcolor{babyblue} 36.5 & \cellcolor{babyblue} 59.6 & 24.0 & \underline{41.7} & \cellcolor{brickred} 23.9 & \cellcolor{brickred} 19.6 & 35.6 & 59.3 & \cellcolor{brickred} 24.4 & \cellcolor{brickred} 51.5 \\
            Llama-3.3 (70B) & 21.6 & 17.6 & \cellcolor{babyblue} 35.8 & \cellcolor{babyblue} 58.8 & 24.1 & \textbf{42.4} & 22.2 & 18.0 & 34.9 & 58.2 & \cellcolor{brickred} 24.6 & \cellcolor{brickred} \underline{51.8} \\
            GPT-4o                  & 26.1 & 21.8 & \cellcolor{babyblue} \textbf{41.8} & \cellcolor{babyblue} \textbf{62.3} & 24.5 & 41.0 & \underline{26.5} & 21.8 & \textbf{40.9} & \textbf{61.6} & \cellcolor{brickred} 24.7 & \cellcolor{brickred} \underline{51.8} \\
            \hline
        \end{tabular}
        \caption{Average coverage and faithfulness scores for the Mixed-Perspective setting. Best model performance on each metric is \textbf{bolded} and second best \underline{underlined}. For each model and metric, significantly higher left than right scores are highlighted in \colorbox{babyblue}{blue}, and significantly higher right than left scores are highlighted in \colorbox{brickred}{red}. Higher scores reflect better performance for all metrics.}
        \label{tab:mo-cov}
    \end{table*}

    \begin{table}[!htbp]
        \centering
        \scriptsize
        \setlength{\tabcolsep}{3pt}
        \begin{tabular}{| P{2.1cm} | P{.65cm} P{.65cm} P{.65cm} | | P{.65cm} P{.65cm} P{.65cm} |}
            \hline
            \multirow{2}{*}{\textbf{Model}} & \multicolumn{3}{c||}{\textbf{Left}} & \multicolumn{3}{c|}{\textbf{Right}} \\
                                    & S$\downarrow$ & O$\uparrow$ & I$\downarrow$ & S$\downarrow$ & O$\uparrow$ & I$\downarrow$ \\
            \hline
            BART$_{CNN/DM}$         &  31.0 &  74.9 & 15.1 &  32.6 &  74.6 &  14.4 \\
            T5                      &  31.1 &  75.6 & 17.8 &  33.7 &  73.3 &  18.7 \\
            Flan-T5                 &  31.0 &  77.2 & 7.9  &  32.1 &  73.5 &   8.0 \\
            Mistral                 &  28.8 &  76.4 & 8.4  &  30.7 &  75.6 &   6.2 \\
            Mixtral                 &  \underline{28.2} &  \underline{76.8} & 10.3 &  29.4 &  \underline{77.0} &   5.9 \\
            Vicuna (7B)             &  29.2 &  76.2 & 9.0  &  30.9 &  75.7 &   6.6 \\
            Vicuna (13B)            &  29.7 &  76.7 & 8.5  &  31.3 &  74.9 &   7.1 \\
            Llama-3.0 (8B)            &  28.9 &  76.7 & 8.7  &  \underline{28.9} &  76.0 &   6.3 \\
            Llama-3.0 (70B)           &  28.8 &  76.1 & \underline{5.8}  &  30.8 &  75.2 &   \underline{5.3} \\
            Llama-3.3 (70B) & 29.1 & 75.8 & \textbf{5.4} & 30.7 & 74.3 & \textbf{4.9} \\
            GPT-4o                  &  \textbf{27.5} &  \textbf{77.1} & 6.8  &  \textbf{28.3} &  \textbf{77.3} &   \underline{5.3} \\
            \Xhline{2pt}
            Human Corr.             & 0.05 & 0.01 & 0.09 & -0.10 & -0.10 & -0.02 \\
            \hline
        \end{tabular}
        \caption{Perspective scores for all models and average sample-level Kendall's Tau correlation between automatic metrics and human judgments (bottom row). Best performance on each metric is \textbf{bolded} and second best \underline{underlined}. \textit{S}: Stance, \textit{O}: Object, \textit{I}: Intensity.}
        \label{tab:mo-rep}
    \end{table}

\section{Results}

    \subsection{Summarization Evaluation}

    \autoref{tab:mo-cov} presents coverage and faithfulness metrics for all models in the mixed-perspective setting. All models outperform the extractive baseline in coverage, emphasizing that abstractive approaches are vital to performing well on this task. Among abstractive approaches, Llama-3.0 (8B) tends to perform best on ROUGE while GPT-4o tends to perform best on \textsc{Bleurt} and BERTScore. Interestingly, while the larger, 13B-parameter Vicuna model performs better than its 7B-parameter variant, the same is not true for Llama-3.0. This suggests that larger models do not always guarantee better mixed-perspective summarization performance. 
    
    Trends in faithfulness metrics differ from that of the coverage metrics. Vicuna models achieve the best performance on \textsc{SummaC} for both perspectives and AlignScore for the right-leaning perspective, whereas Llama-3.3 (70B) is best on AlignScore for the left perspective. Overall, however, all models score far lower on faithfulness metrics than the extractive baseline, suggesting improving model faithfulness to the intended perspective for 
    this task is an important issue to address in future work. 
    
    Between perspectives, BLEURT and \textsc{BERTScore} are significantly higher for left-leaning summaries across most models, while \textsc{SummaC} and AlignScore are significantly higher for most right-leaning summaries. Notably, all four metrics are model-based as opposed to ROUGE metrics which exhibit few significant differences between left and right-leaning summaries. Because the differences are consistent across most approaches, and prior works have uncovered political biases in models similar to those underlying each metric (e.g., \citealt{feng-pretraining}), it suggests that these metrics may also suffer from political biases. 

\subsection{Perspective Evaluation}

    \begin{table}[!htbp]
        \centering
        \small
        \begin{tabular}{| P{1cm} | P{.4cm} P{.4cm} P{.4cm} P{.4cm} P{.4cm} | P{.5cm} P{.5cm} |}
        \hline
            \textbf{Model} & S & O & I & C & R & Avg. Len & Avg. Ext \\
            \hline
            Ref.      & 2.9 & 3.2 & 2.9 & 2.9 & 1.8 & 19.7 & 0.68 \\
            \hline
            Mistral   & 2.4 & 3.2 & 2.5 & 2.5 & 1.9 & 48.6 & 0.84 \\
            Mixtral   & 3.0 & 3.4 & 3.0 & 3.0 & 2.6 & 46.2 & 0.72 \\
            GPT-4o    & \textbf{3.2} & \textbf{3.5} & \textbf{3.2} & \textbf{3.2} & \textbf{2.9} & 39.7 & 0.66 \\
            \hline
        \end{tabular}
        \caption{Average human judgments of each perspective dimension with average length and extraction coverage over all inputs. Best scores are \textbf{bolded}. Higher scores reflect better performance for all metrics. Length is measured in Llama-3 tokens. \textit{S}: Stance, \textit{O}: Object, \textit{I}: Intensity, \textit{C}: Consensus, \textit{R}: Reasoning.}
        \label{tab:human-judge}
    \end{table}

        \begin{table*}[!htbp]
        \centering
        \scriptsize
        \setlength{\tabcolsep}{4pt}
        \begin{subtable}[h]{0.54\textwidth}
            \centering
            \begin{tabular}{| P{1.5cm} | P{6.5cm}|}
                \hline
                \multicolumn{2}{|c|}{\textbf{2020 Census Battle}}  \\
                \hline
                Model & Summary  \\
                \hline
                \textit{Reference} & The left argues that the new evidence makes clear the GOP’s true motives, and condemns the citizenship question as undermining the purpose of the census.\\
                \Xhline{2pt}
                Mistral & The left argues that the {\color{blue} Trump administration's efforts to include a citizenship question on the 2020 census} are a {\color{blue} partisan power grab} that undermines the accuracy and fairness of the census, and that the real goal is to {\color{blue} rig a constitutional mandate.}\\
                \hline
                Mixtral & The left perceives the Trump administration's attempt {\color{blue} to include a citizenship question on the 2020 census as a partisan power grab} aimed at benefiting {\color{orange} Republicans and non-Hispanic Whites}, potentially leading to an {\color{purple} undercount of millions of people} and {\color{blue} an unfair redistricting cycle.}\\
                \hline
                GPT-4o & The left views the {\color{blue} Trump administration's push to include a citizenship question on the 2020 census as a partisan maneuver} {\color{orange} aimed at disenfranchising minority voters and skewing political power in favor of Republicans and non-Hispanic Whites.}\\                
                \hline
            \end{tabular}
            \caption{Models often extract different information from individual sources rather than synthesizing the inputs into a single perspective summary.}
            \label{tab:summ-ex1a}
        \end{subtable}
        \hfill
        \begin{subtable}[h]{0.42\textwidth}
            \centering
            \begin{tabular}{| P{1.5cm} | P{4.5cm} |}
                \hline
                \multicolumn{2}{|c|}{\textbf{Amy Coney Barret Confirmed}}  \\
                \hline
                Model & Summary \\
                \hline
                \textit{Reference} & The right applauds Barrett’s confirmation.	\\
                \Xhline{2pt}
                Mistral & The right: {\color{blue} Republicans did not break any norms} in confirming Amy Coney Barrett to the Supreme Court.\\
                \hline
                Mixtral & The right-leaning political perspective within these texts emphasizes that {\color{blue} Republicans have not broken any norms or rules in confirming their nominees to the Supreme Court, and that the only rule is having the power to actuate a party's political will.}\\
                \hline
                GPT-4o & The right argues that {\color{blue} Republicans have adhered to historical norms and constitutional principles in their judicial nominations}, {\color{orange} emphasizing judicial restraint and originalism}, while Democrats have engaged in politicizing the process and undermining these norms. \\
                \hline
            \end{tabular}
            \caption{Models tend to extract from the same few input documents.}
            \label{tab:summ-ex1b}
        \end{subtable}
        \caption{Example perspective summary generations for Mistral, Mixtral, and GPT-4o. Distinct colors signify near paraphrases or exact quotes from a particular document identified manually by the authors.}
        \label{tab:summ-ex1}
    \end{table*}

    Automatic perspective metrics are included in \autoref{tab:mo-rep}, and human perspective judgments of Mistral, Mixtral, GPT-4o, and reference summaries are included in \autoref{tab:human-judge} (breakdown of judgment scores and correlations with automatic metrics are included in \autoref{app:judgments} and \autoref{app:summ-corr}). 
    GPT-4o summaries tend to score highest on both automatic and human scores, even exceeding reference summary judgments. While the references are similar in extractiveness to GPT-4o, the summaries are significantly shorter which may make them difficult to accurately compare. 

    \begin{table}[!htbp]
        \centering
        \scriptsize
        \begin{tabular}{| P{.5cm} | P{4cm}| P{.6cm} P{.8cm} |}
            \hline
            \multicolumn{4}{|c|}{\textbf{Attorney General William Barr Testifies}}  \\
            \hline
            Side & Summary & Align & SummaC \\
            \hline
            Left & The left believes that Attorney General Barr's handling of the Mueller report was a deliberate attempt to mislead the public and protect President Trump from accountability, and that Congress must take action to hold Trump accountable for his abuses of power. & 0.8 & 22.2 \\
            \hline
            Right & The right believes that Attorney General Barr acted appropriately in releasing a summary of the Mueller report, and that Democrats are unfairly attacking him for not spinning the report in their favor, while also downplaying the fact that Mueller's report did not find collusion and did not recommend obstruction charges. & \textbf{94.4} & \textbf{24.4} \\
            \hline
        \end{tabular}
        \caption{Example pair of left and right-leaning summaries generated by Llama-3.0 (70B) with AlignScore and \textsc{SummaC} scores.}
        \label{tab:summ-ex2}
    \end{table}
    
    Compared to human judgments, the automatic metrics for Stance, Object, and Intensity largely rank the three models in the same order, although there is little sample-level correlation with human judgments.
    While the metrics may capture system-level differences (i.e., between GPT-4o and Mistral), they often fail to distinguish individual summaries of similar quality (i.e., summaries by the same model). Improving metrics for measuring similarity between perspective dimensions and exploring alternative methods for measuring perspective dimensions are vital areas for future work.

\section{What Drives Model Performance?}

    In this section, we now turn to investigating what factors may be driving performance or present obstacles to high quality summaries. We first qualitatively analyze example summaries, and then analyze the relationship between input features and models' extraction behavior. In analyzing model extraction behavior, we specifically measure the extent to which models extract from each input document, and what qualities of input documents are associated with higher rates of extraction.

    \subsection{Qualitative Evaluation}
    \label{sec:qual-eval}

    Examples summaries generated by Mistral, Mixtral, and GPT-4o are shown in \autoref{tab:summ-ex1}. In these examples, models appear to attend to the correct perspective within input documents, suggesting that models may have sufficient knowledge of left and right-leaning political beliefs and rhetoric to distinguish inputs in some cases. In summarizing the perspectives, however, models consistently tend to rely primarily on specific information 
    present in 
    a small subset of input documents with little abstractive synthesis of the information (\autoref{tab:summ-ex1a}). 
    Additionally, models consistently extract or paraphrase from the same documents across many examples, such as in \autoref{tab:summ-ex1b}.
    This consistent behavior 
    suggests that qualities of the input documents themselves may drive models' tendency to extract from them, which we investigate in \autoref{sec:extract-analysis}.

        \begin{figure*}[!htbp]
            \centering
            \includegraphics[width=0.95\linewidth]{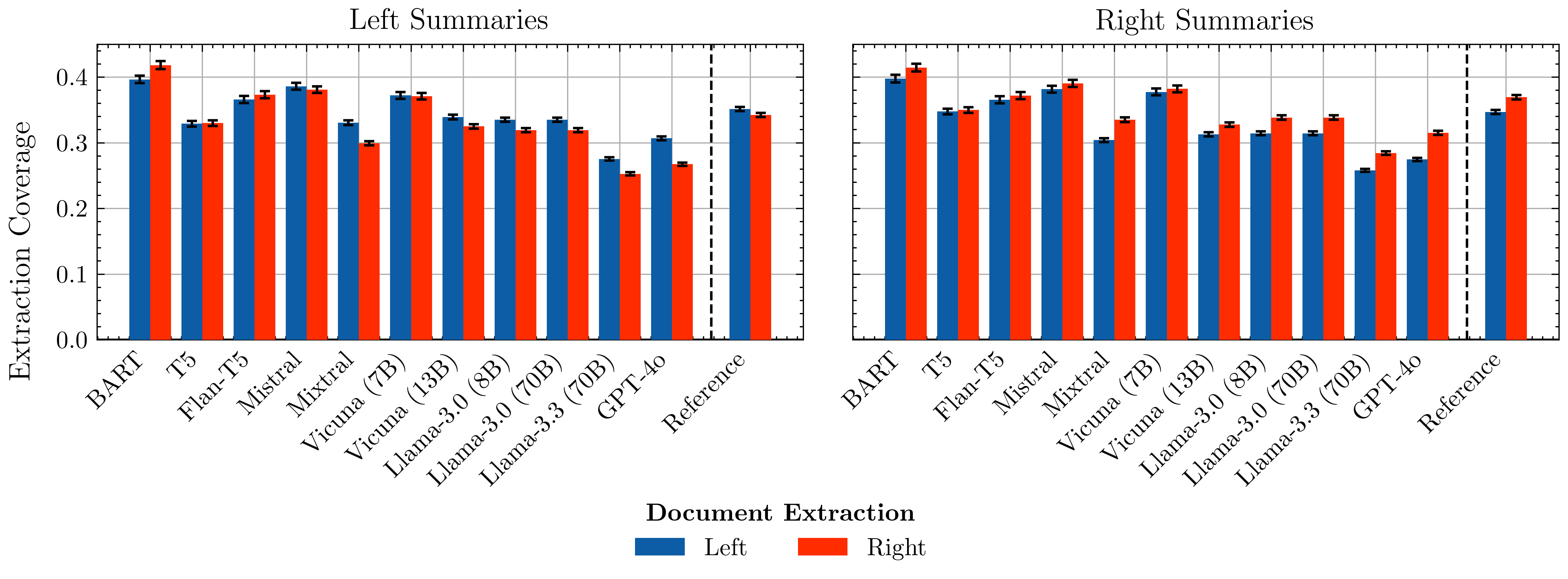}
            \caption{Average extraction coverage of left and right-leaning source documents for each model shown separately for left and right-leaning summaries. Only the top 4 source documents, 2 for each perspective, are considered. Error bars reflect 95\% confidence intervals.}
            \label{fig:side-extract}
        \end{figure*}

    Furthermore, \autoref{tab:summ-ex2} shows a selected pair of left and right summary examples where the right summary faithfulness metrics are notably larger than that of the left. In this case, neither summary contains noticeable faithfulness errors relative to the source documents, but there is a 2 point difference in SummaC scores and a 93 point difference in AlignScore both favoring the right-leaning summary. Throughout the dataset, cases resembling \autoref{tab:summ-ex2} where perspective summaries are faithful to similar extents support that neural metrics may suffer from political biases or spurious correlations beyond faithfulness of the input.

    \subsection{Extraction Analyses}
    \label{sec:extract-analysis}

        To further understand model performance on this task, we analyze extraction behavior with respect to different qualities of the input documents: political perspective, position in the input, length, and use of 
        arousing terms. We use \textit{extractive fragment coverage} to quantify this behavior, as defined by \citet{grusky-newsroom}. 
        In analyses between the sets of left and right-leaning input documents,
        we hypothesize that models with better
        performance tend to extract more from the intended political perspective relative to the alternative perspective. Accordingly, we expect less performant models to extract similarly from both perspectives or more from the unintended perspective. For comparisons among documents sharing a common political perspective, we evaluate three additional hypotheses illustrating potentially undesirable behaviors: models extract more information from documents that \textbf{(1)} appear earlier in the input (position bias), \textbf{(2)} are longer (length bias), and \textbf{(3)} that contain more arousing language (arousal bias).  

        \paragraph{Document Perspective.} As the Mixed-Perspective setting requires models to disentangle perspective among input documents, we first analyze whether models tend to extract more tokens from source documents with the correct political slant. To investigate this relationship, \autoref{fig:side-extract} presents the extraction coverage scores for each combination of summary and source document perspective (e.g., \textit{left/right} perspective summary and \textit{left/right}-leaning source documents). For both perspectives, performant models tend to extract from source documents portraying the intended perspective (blue bars in the "Left Summaries" plot, and red bars in "Right Summaries") significantly more often than the alternative perspective. The less performant T5, Flan-T5, and Mistral models, however, present insignificant differences in the extent that they extract from the intended perspective with respect to the other perspective.
        In other words, these models extract similarly from left and right-leaning inputs regardless of the intended perspective of the summary. This suggests that poor performance of the models may be due to the inability to distinguish political leanings among input documents. Uniquely, BART consistently extracts significantly more from right-leaning documents than left-leaning, regardless of the intended perspective. 
        Therefore, \textbf{the largest and best performing models (e.g., GPT-4o, Llama-3.3, Llama-3.0, etc.) appear to be sufficiently capable of distinguishing perspectives in inputs.} 

        \begin{figure*}[!htbp]
            \centering
            \includegraphics[width=1.0\linewidth]{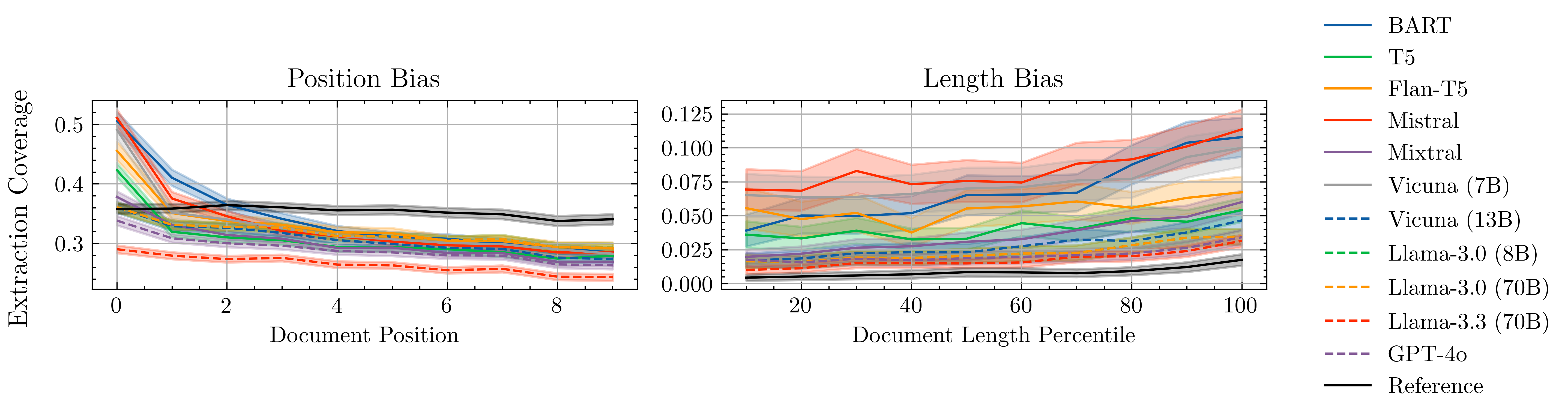}
            \caption{Average extraction coverage of source documents by position in the input (left) and length of input document (right) for each model. For input document length, percentiles buckets are shown rather than raw values. Extraction coverage for length bias analysis only considers terms that appear in a single input document. Smaller numbers reflect earlier input positions. Shaded regions reflect 95\% confidence intervals at each document position.}
            \label{fig:pos-length-bias}
        \end{figure*}
    
        \begin{figure}[!htbp]
            \centering
            \includegraphics[width=0.98\linewidth]{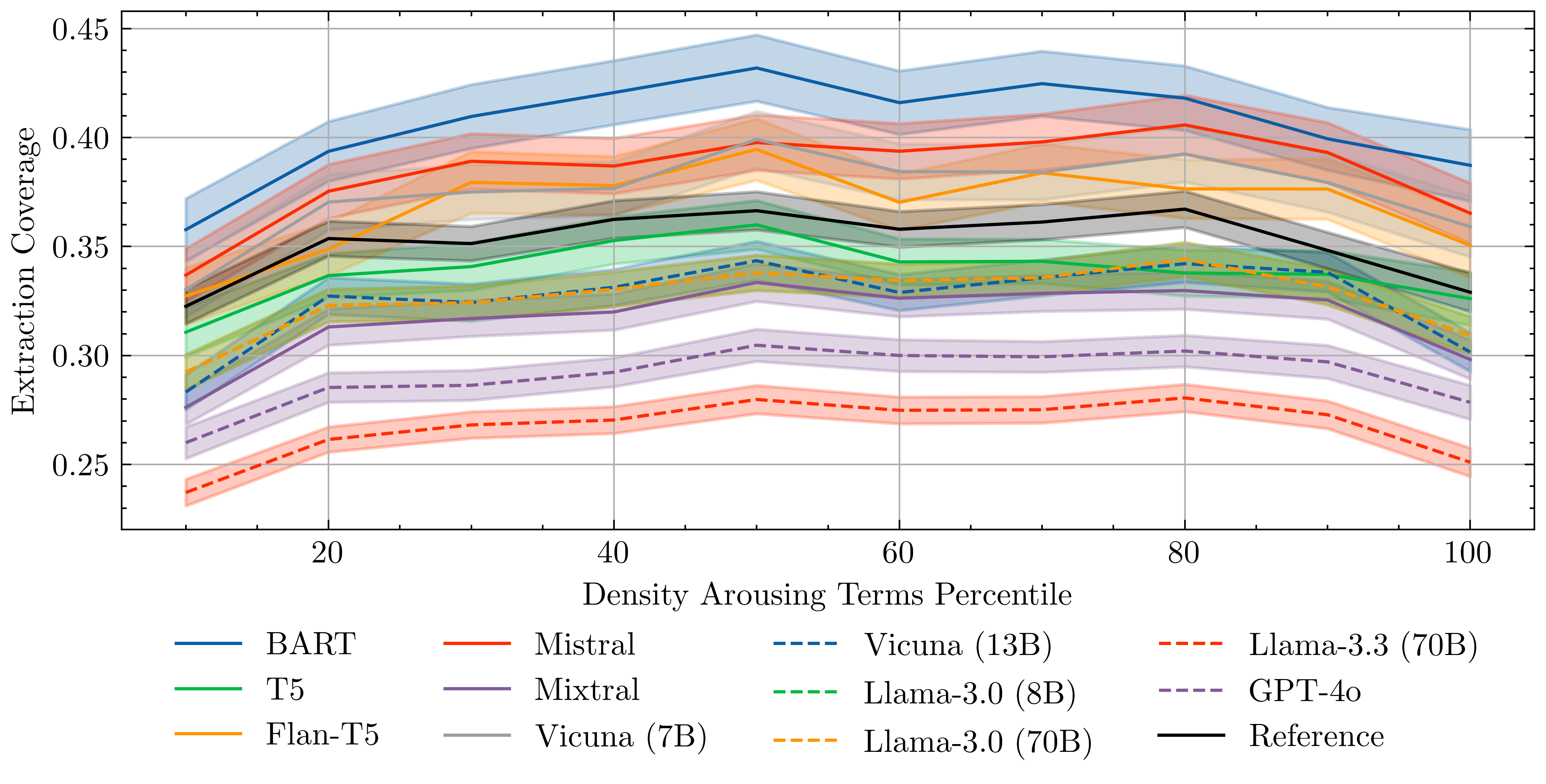}
            \caption{Average extraction coverage of source documents for each model by density of arousing terms in the source document as percentiles.}
            \label{fig:narous-extract}
        \end{figure}

        \paragraph{Document Position.} Position and lead biases in summarization systems are well-documented phenomena \cite{chhabra-revisiting, olabisi-position, jung-earlier}. Following these works, \autoref{fig:pos-length-bias} presents the extraction coverage scores across document positions for each model. As in prior research, 
        we find a near monotonically decreasing relationship between extraction from a document and its position in the input, with all models drawing significantly more terms from the first document compared to those that follow. Among models, Llama-3.3 (70B) and GPT-4o appear to be most robust to document position, indicated by the flatter curves, but still exhibit more bias than human-written references. Surprisingly, however, Mistral appears to be most susceptible to position biases and scores higher on extraction coverage than the smaller BART, T5, and Flan-T5 models. Overall, \textbf{all models suffer from position biases and struggle to incorporate documents that appear later in inputs}.

        \paragraph{Document Length.} In addition to position biases, we examine how extraction coverage varies with length of the input documents in \autoref{fig:pos-length-bias}. As longer documents are more likely to contain 
        terms shared with other documents, in this setting, we modify the extraction coverage metric to only consider terms unique to an individual input document. In doing so, we more precisely measure to what extent models extract from a particular document in relation to document length.
        In this case, we see a slight positive trend in extraction of unique terms with document length percentile among models compared to the human-written references\footnote{For all models excluding Flan-T5, the extraction coverage of the top percentile of document lengths is significantly larger ($p\leq 0.05$) than that of the bottom percentile}. \textbf{We find that models tend to extract more unique terms from longer documents}, but the impact appears less pronounced than with document position. 

        \paragraph{Document Arousal.} Finally, we consider how the \textit{language} of documents may influence models' content selection. In particular, we consider the relationship between the density of highly arousing terms in an input document and the extent to which the document is represented in the summary, shown in \autoref{fig:narous-extract}.
        In contrast to the position and length analyses, the relationship between extraction coverage and arousal of an input document is not 
        consistent across densities of arousing terms. Instead, all models appear to extract less from both non-arousing and highly arousing documents, and to a lesser extent, reference summaries exhibit this behavior as well. This may be because models and humans struggle to incorporate underrepresented perspectives
        (i.e., documents that present a perspective that diverges further from the other documents in the input). As with other analyses, the largest models appear to be most robust to this trend though still present a similar behavior. Therefore, contrary to our initial hypothesis, we conclude that \textbf{models tend to avoid extracting from documents that use extremely many \textit{or} extremely few arousing terms} relative to other input documents.

\section{Related Work}

    \paragraph{Opinions and Summarization. } Mining and summarizing opinions have long been an interest to the NLP community. Approaches to opinion summarization, typically a multi-document task, have included two-stage extraction-abstraction models \cite{op-digest,twostep-opsum}, pointer-generator networks \cite{copycat,point-gen2}, and deep controllable language models \cite{control-op-summ}. Summarization in other domains, such as news summarization, can also be inherently opinionated. In these cases,
    prior work has shown that LLMs can fail to fairly represent the diversity of perspectives among inputs \cite{zhang-fair, huang-bias} or account for biases within input texts \cite{tay-review}. Furthermore, recent work has found that such behaviors are realized differently when multiple dialects are represented \cite{olabisi-divsumm, olabisi-position}.
    In contrast to these works, we focus on a multi-document, multi-summary task with the aim of summarizing political perspectives independently.

    \paragraph{Political Ideologies and LMs.}
    Recent work has increasingly investigated the political biases and leanings of language models. \citet{santurkar-opinions} identify ideologies that language models best align with, while others such as \citet{feng-pretraining} evaluate the impact of partisan pretraining data on downstream stream task performance.
    Particularly in summarization, evaluated models have been found to hold name-nationality biases \cite{ladhak-name-nationality}, journalistic framing biases \cite{lee-neus}, geopolitical biases \cite{li-geopolitical}, as well as biases related to perspectives in input documents \cite{rajan-viewpoints}.
    While our work similarly focuses on a political summarization task, our work uniquely evaluates models' abilities to summarize individual political perspectives rather than a single bias-free or composite summary.
    
    \paragraph{Contrastive, Comparative, and Neutral Summarization. }
    To combat the aforementioned biases in summarization, some have worked toward generating neutral summaries free of journalistic framing bias \cite{pol-bias-rein, bang-mitigating}, as well as aligning the stance of the input documents and generated summaries \cite{liu-p3}. Rather than improving the biases of a single summary, a separate line of inquiry has aimed to generate contrasting summaries. Contrastive Opinion Summarization (COS) was originally dominated by extractive approaches \cite{cos-task}, selecting a pair of inputs conditioned on each other to be the most contrastive and representative. More recently, work has overcome extractive approach limitations and developed abstractive systems \cite{comparative-opinion-summ}. These and other abstractive approaches typically rely on knowledge of the perspectives represented by each input. While in our work, we similarly generate independent summary pairs, we instead evaluate how well LLMs' can summarize \textit{without} prior knowledge of input perspectives.

\section{Conclusion}

    In this work, we introduce a dataset, \textsc{PoliSum}, for generating a summary of the left-perspective and a summary of the right-perspective given a set of politically opinionated texts. For this task, we also introduce a framework for evaluating perspective summaries through both human evaluation and automatic metrics. We benchmark current models of varying sizes and architectures on this task, analyzing what behaviors drive performance through their tendency to extract from particular documents.
    
    Models that are capable of accurately and faithfully summarizing multiple perspectives can help alleviate polarization and misunderstandings through exposing others to alternative perspectives than those they hold. While we find that larger and more recent models like GPT-4o can perform well on this task, we also identify problems with faithfulness common among all models evaluated. Additionally, biases due to document position, document length, and use of arousing terms may hinder performance. We recommend future work focus on mitigating these issues while also progressing toward generalization to arbitrary perspectives.

\section{Limitations}

We adopt existing coverage and faithfulness metrics for traditional summarization task as part of our automatic evaluation of summaries. These metrics, however, may not be equally appropriate to the proposed perspective summarization setting. Through additional automatic and human evaluation, we aim to capture aspects of performance specific to summarizing perspectives to complement these results. We encourage future work to investigate the existing coverage and faithfulness metrics on this task, as well as develop novel metrics to ensure accurate measures of performance.

Additionally, while we focus on capturing varied political perspectives, we acknowledge that the left-right dichotomy may not fully capture the broad range of existing political views \cite{feldman-ideology}. We restrict our initial evaluation to left and right-leaning perspectives due to data availability, but the task formulation and proposed evaluation framework can be applied to arbitrary sets of perspectives. We encourage future work to evaluate and develop approaches for summarizing a broader range of points-of-view in order to represent such perspective diversity.

Finally, some of the models evaluated may lack appropriate training on recent political issues to enable distinguishing the perspectives presented in source documents. Data in PoliSum covers a time span from September 2018 to January 2024, but the last article of the CNN/DM dataset, for example, was written in April 2015 \cite{hermann-cnndm}. This lack of recent knowledge may introduce additional challenges to summarizing source passages covering topics such as COVID-19 or recent elections. Future work might explore topic-invariant approaches to summarizing perspectives that are able to adapt to current events.

\section{Ethics Statement}

Hallucinations and factual inconsistencies in language models have received widespread attention due to the risks to safety and potential for misinformation \cite{huang-factual, ji-hallucinate}. Though the intent of our work is to generate distinct summaries of perspectives in order to better represent each, misrepresentations due to these hallucinations could lead to misunderstandings of opposing perspectives and the accompanying impacts on polarization \cite{braley-democracy, lees-polarization}. Additionally, language models that may elevate certain perspectives over others can further contribute to these impacts. We acknowledge that the collected data could be misused in order to intentionally induce these behaviors in language models toward a particular political perspective. In this work, we avoid conducting experiments involving finetuning models and strictly use the collected data to evaluate summaries in order to highlight such weaknesses in existing models. All data in this work is used with permission from The Flip Side and human evaluations included in this work are conducted under an approved IRB protocol.

\section*{Acknowledgments}

This work was supported in part by grant IIS-2106666 from the National Science Foundation, the Knight First Amendment Institute at Columbia University, National Science Foundation Graduate Research Fellowship DGE-2036197, the Columbia University Provost Diversity Fellowship, and the Columbia School of Engineering and Applied Sciences Presidential Fellowship. Any opinion, findings, and conclusions or recommendations expressed in this material are those of the authors and do not necessarily reflect the views of the National Science Foundation or the Knight First Amendment Institute. We thank the anonymous reviewers and Smaranda Muresan for providing feedback on an earlier draft of the work.

\bibliography{custom}

\appendix

\section{Zero-Shot Baselines}
\label{app:zs-baselines}

    \begin{table}[!htbp]
        \centering
        \small
        \begin{tabular}{| P{2cm} | P{4cm} |}
            \hline
            Models & Prompt \\
            \hline
            T5, Flan-T5 & ``summarize the [left,right]-leaning political perspective: \{PASSAGES\}'' \\
            \hline
            Mistral, Mixtral, Vicuna, Llama-3.0, Llama-3.3, GPT-4o & ``Produce a short, single-sentence summary of the [left,right]-leaning political perspective within the following texts. Respond only with the summary beginning with "The [left,right]": \{PASSAGES\}'' \\
            \hline
        \end{tabular}
        \caption{Summarization prompts provided to each model. "left" and "right" are included in the prompt depending on the intended political perspective. }
        \label{tab:prompts}
    \end{table}

    Because it is finetuned on CNN/DM, inference with BART is done without a prompt. BART summaries, however, are forced to begin with either "The left" or "The right" depending on the intended perspective to enforce the task format. Inference with T5 and Flan-T5 is conducted using the first prompt in \autoref{tab:prompts}. Summaries for all three models are generated with the default transformers package hyperparameters and greedy decoding. 

\section{Prompting Baselines}
\label{app:prompt-baselines}

    For larger LLMs, we use the second prompt in \autoref{tab:prompts}, generating each perspective summary separately. Larger models are accessed via OpenAI (GPT-4o) and Fireworks AI\footnote{\url{https://fireworks.ai}} (Llama-3.3-70B, Llama-3.0-70B, Vicuna-13B, and Mixtral-8x22B) API's. In all cases, we use a temperature of 0 to ensure deterministic outputs and greedy decoding. Smaller model (Mistral-7B, Llama-3.0-8B, and Vicuna-7B) inference is run with bfloat16 precision, also using greedy decoding.

\section{Data Statement}
\label{app:data-statement}

    Details characterizing the collected data are included in the following data statement.

    \subsection{Curation Rationale}

        The data was collected in order to evaluate models in summarizing political perspectives on controversial issues provided passages from news articles and op-eds. All data comes from The Flip Side, a site which publishes summaries of left and right-leaning perspectives on current issues with accompanying passages from other news sources. Each sample includes a headline describing the controversial issue, a set of left and right-leaning source passages, and pair of one-sentence left and right perspective summaries.

    \subsection{Language Variety}

        All texts are in English (en-US) and reflect a formal variety of English commonly found in news articles.

    \subsection{Speech Situation}

        All source passages and summaries are originally written texts. As the data is primarily comprised of passages from news articles and op-eds, the texts were intended for a broad American audience, written asynchronously, and are likely edited repeatedly before publishing. All data was originally published between September 2018 and January 2024.

    \subsection{Text Characteristics}

        A majority of the texts in the dataset reflect politically controversial issues, including topics such as elections, policy, political figures, and cultural phenomena. As the task focuses on summarizing political perspectives, the texts also heavily reflect opinions of the original authors in the case of the source passages, or a general left or right-leaning perspective in the case of the reference summaries.

\section{Evaluation Metrics Implementation}
\label{app:metric-details}
    
    We evaluate models using existing implementations of coverage, consistency, and factuality metrics. \textsc{Rouge} is calculated using the \textsc{Rouge}-score python package \cite{lin-rouge}. \textsc{BERTScore} is calculated using the Torchmetrics implementation\footnote{\url{https://github.com/Lightning-AI/torchmetrics}}. All other metrics are calculated using the implementations by the original authors for each: \textsc{Bleurt} \cite{sellam-bleurt}, \textsc{SummaC} \cite{laban-summac}, and \textsc{AlignScore} \cite{zha-alignscore}. All coverage, faithfulness, and automatic metrics for perspective dimensions (Stance, Object, and Intensity) are scaled to a 0-100 range.

\section{Single-Perspective Setting}
    \label{app:single-persp}

    \subsection{Methods}
    
        In the Single-Perspective Setting, models are evaluated similarly to traditional opinion summarization. Models are provided with passages from a single perspective only (i.e., left or right-leaning passages), and asked to generate a summary for the same perspective. Therefore, in contrast to the Mixed-Perspective Setting, models are not expected to distinguish input passages.

    \subsection{Results}

        Automatic coverage and faithfulness metrics for the Single-Perspective Setting are shown in \autoref{tab:sp-cov}, while automatic perspective dimension scores are shown in \autoref{tab:sp-rep}.
    
    \begin{table*}[!ht]
        \centering
        \scriptsize
        \begin{tabular}{| P{2cm} | P{.6cm} P{.6cm} P{.9cm} P{.6cm} | P{.8cm} P{.6cm} | | P{.6cm} P{.6cm} P{.9cm} P{.6cm} | P{.8cm} P{.6cm} |}
            \hline
            \multirow{3}{*}{\textbf{Model}} & \multicolumn{6}{c||}{\textbf{Left}} & \multicolumn{6}{c|}{\textbf{Right}} \\
                                    & \multicolumn{4}{c|}{Coverage} & \multicolumn{2}{c||}{Faithfulness} & \multicolumn{4}{c|}{Coverage} & \multicolumn{2}{c|}{Faithfulness} \\
                                    & R1   & RL   & \textsc{Bleurt} & BERT& \textsc{SummaC} & Align & R1   & RL    & \textsc{Bleurt} & BERT & \textsc{SummaC} & Align \\
            \hline
            BART$_{CNN/DM}$         & 25.4 & \underline{22.8} & \cellcolor{babyblue} 30.7 & \cellcolor{babyblue} \underline{61.6} & 25.6 & 30.3 & 25.5 & 22.9 & 29.9 & 59.9 & \cellcolor{brickred} 27.0 & \cellcolor{brickred} 38.4 \\
            T5                      & 24.6 & 22.3 & 30.1 & \cellcolor{babyblue} 60.8 & 23.5 &  16.3 & \cellcolor{brickred} 25.5 & \cellcolor{brickred} \underline{23.1} & \cellcolor{brickred} 31.2 & \underline{60.3} & \cellcolor{brickred} 23.8 & 16.8 \\
            Flan-T5                 & 23.0 & 20.2 & 28.9 & \cellcolor{babyblue} 59.0 & \underline{27.3} & 42.4 & \cellcolor{brickred} 23.9 & \cellcolor{brickred} 21.2 & 28.6 & 58.6 & \underline{27.3} & 44.6 \\
            Mistral                 & \underline{26.1} & 22.2 & \cellcolor{babyblue} 35.9 & \cellcolor{babyblue} 61.1 & 26.3 & \textbf{52.1} & 26.0 & 22.2 & 34.7 & 59.8 & \cellcolor{brickred} 27.2 & 52.7 \\
            Mixtral                 & \underline{26.1} & 22.2 & 38.6 & 60.4 & 26.6 & 50.7 & \cellcolor{brickred} 27.2 & \cellcolor{brickred} 23.0 & 38.0 & 60.2 & \cellcolor{brickred} 26.9 & \textbf{61.4} \\
            Vicuna (7B)             & 24.9 & 21.5 & 33.9 & \cellcolor{babyblue} 59.5 & \textbf{29.6} & 47.3 & \cellcolor{brickred} 26.0 & \cellcolor{brickred} 22.4 & 33.7 & 58.6 & \cellcolor{brickred} \textbf{31.7} & \cellcolor{brickred} 58.0 \\
            Vicuna (13B)            & 26.0 & 22.4 & \cellcolor{babyblue} 38.2 & \cellcolor{babyblue} 60.3 & 26.5 & \underline{51.6} & 26.3 & 22.8 & 37.0 & 59.4 & 26.3 & \cellcolor{brickred} \underline{60.3} \\
            Llama-3.0 (8B)  & \textbf{27.3} & \textbf{23.0} & \underline{39.2} & \cellcolor{babyblue} 60.1 & 25.5 & 46.0 & \cellcolor{brickred} \textbf{28.0} & \cellcolor{brickred} \textbf{23.9} & \underline{38.9} & 59.4 & \cellcolor{brickred} 25.9 & \cellcolor{brickred} 54.3 \\
            Llama-3.0 (70B) & 23.8 & 19.6 & \cellcolor{babyblue} 37.7 & \cellcolor{babyblue} 60.2 & 24.2 & 39.0 & \cellcolor{brickred} 25.0 & \cellcolor{brickred} 20.5 & 36.7 & 59.8 & 24.4 & \cellcolor{brickred} 52.6 \\
            Llama-3.3 (70B)         & 22.2 & 18.1 & \cellcolor{babyblue} 37.0 & \cellcolor{babyblue} 59.1 & 24.5 & 42.1 & \cellcolor{brickred} 22.8 & \cellcolor{brickred} 18.6 & 35.8 & 58.5 & 24.7 & \cellcolor{brickred} 53.9 \\
            GPT-4o                  & 25.6 & 21.8 & \textbf{41.3} & \cellcolor{babyblue} \textbf{62.1} & 25.1 & 44.2 & \cellcolor{brickred} \underline{27.0} & \cellcolor{brickred} \underline{23.1} & \textbf{40.8} & \textbf{61.5} & \cellcolor{brickred} 25.5 & \cellcolor{brickred} 54.2 \\
            \hline
        \end{tabular}
        \caption{Average coverage and faithfulness scores for the Single-Perspective setting. Best performance on each metric is \textbf{bolded} and second best \underline{underlined}. 
        For each model and metric, significantly higher left than right scores are highlighted in \colorbox{babyblue}{blue}, and significantly higher right than left scores are highlighted in \colorbox{brickred}{red}. 
        Higher scores reflect better performance for all metrics.}
        \label{tab:sp-cov}
    \end{table*}

    \begin{table}[!htbp]
        \centering
        \scriptsize
        \setlength{\tabcolsep}{3pt}
        \begin{tabular}{| P{2.1cm} | P{.65cm} P{.65cm} P{.65cm} | | P{.65cm} P{.65cm} P{.65cm} |}
            \hline
            \multirow{2}{*}{\textbf{Model}} & \multicolumn{3}{c||}{\textbf{Left}} & \multicolumn{3}{c|}{\textbf{Right}} \\
                                    & S$\downarrow$ & O$\uparrow$ & I$\downarrow$ & S$\downarrow$ & O$\uparrow$ & I$\downarrow$ \\
            \hline
            BART$_{CNN/DM}$         & 30.3 & 74.7 & 15.1 & 31.7 & 74.4 & 14.0 \\
            T5                      & 31.3 & 76.0 & 17.9 & 32.5 & 73.5 & 19.4 \\
            Flan-T5                 & 31.7 & \underline{77.2} & 7.9  & 32.6 & 73.8 & 8.0 \\
            Mistral                 & 28.5 & \textbf{77.4} & 8.1  & 30.0 & 75.5 & 6.8 \\
            Mixtral                 & 27.4 & 76.9 & 9.6  & 28.9 & \textbf{77.2} & 5.8 \\
            Vicuna (7B)             & 29.5 & 76.2 & 9.1  & 29.1 & 75.5 & 6.5 \\
            Vicuna (13B)            & \underline{27.2} & 77.0 & 7.2  & 31.0 & 76.3 & 6.2 \\
            Llama-3.0 (8B)  & \textbf{26.7} & 77.0 & 7.8  & \textbf{28.5} & \underline{77.1} & 6.2 \\
            Llama-3.0 (70B) & 29.7 & 77.1 & \underline{6.1}  & 30.2 & 77.0 & \underline{5.1} \\
            Llama-3.3 (70B)         & 27.6 & 75.3 & \textbf{5.1}  & 30.8 & 73.3 & \textbf{4.8} \\
            GPT-4o                  & 27.6 & \textbf{77.4} & 6.9  & \underline{28.6} & \underline{77.1} & 5.5 \\
            \hline
        \end{tabular}
        \caption{Perspective scores for all models in the Single-Perspective setting. 
        Best performance on each metric is \textbf{bolded} and second best \underline{underlined}. 
        \textit{S}: Stance, \textit{O}: Object, \textit{I}: Intensity.}
        \label{tab:sp-rep}
    \end{table}

\section{Human Judgments}
\label{app:judgments}

\subsection{Annotator Demographics}

    Human judgments are conducted with 6 undergraduate students pursuing a BS or BA in Political Science, with self-reported knowledge of the common beliefs of conservatives and liberals. As such, they have the required expertise to judge perspectives conveyed in summaries. Demographics of the recruited annotators are shown in \autoref{tab:annot-demo}. 

    \begin{table}[!htbp]
        \centering
        \small
        \begin{tabular}{| P{1.5cm} P{4cm} | P{.8cm} |}
            \hline
            Demographic Variable & Group & \% \\
            \hline
            \multirow{2}{1.5cm}{Gender} & Male & 66.6\% \\
            & Female & 33.3\% \\ 
            \hline
            \multirow{7}{1.5cm}{Political Ideology} & Extremely Liberal & 16.6\% \\
            & Liberal & 33.3\% \\
            & Slightly Liberal & 33.3\% \\
            & Moderate & 16.6\% \\
            & Slightly Conservative & 0\% \\
            & Conservative & 0\% \\
            & Extremely Conservative & 0\% \\
            \hline
            \multirow{8}{1.5cm}{Race/ Ethnicity} & American Indian, Alaskan Native, and/or Indigenous & 16.6\% \\
            & Arab American, Middle Eastern, or North African & 0\% \\
            & Asian or Asian American & 0\% \\
            & Black or African American & 16.6\% \\
            & Latino/a/x or Spanish Origin & 16.6\% \\
            & Native Hawaiin or Pacific Islander & 0\% \\
            & Southeast Asian & 0\% \\
            & White/European American & 66.6\% \\
            & Other & 16.6\% \\
            \hline
        \end{tabular}
        \caption{Demographic statistics of the 6 annotators.}
        \label{tab:annot-demo}
    \end{table}

\subsection{Human Judgment Correlations by Model}
\label{app:judge-corr}

    Correlation with human judgments by model is shown in \autoref{tab:corr-brkdwn}. As in the overall sample-level results, correlations with human judgments are largely poor and non-significant. For GPT-4o, however, the automatic metric for Object is significant and weakly negatively correlated with human judgments.

    \begin{table}[!htbp]
        \centering
        \begin{tabular}{| c | c c c |}
            \hline
            Model   & S     & O     & I \\
            \hline
            Mistral & 0.04  & 0.04  & 0.03 \\
            Mixtral & -0.03 & 0.01  & -0.08 \\
            GPT-4o  & -0.06 & -0.19* & 0.02 \\
            \hline
        \end{tabular}
        \caption{Sample-level correlation with human judgments for each perspective dimension by model. Significant correlations are marked with * ($p\leq.05$).}
        \label{tab:corr-brkdwn}
    \end{table}

\subsection{Human Judgments by Perspective}
\label{app:judge-breakdown}
    
    Human judgments of perspective dimensions for both the left and right perspectives are shown in \autoref{tab:human-judge-full}. In both perspectives, GPT-4o summaries are scored highest for all dimensions. While some average scores appear to be slightly higher for the left perspective, no differences between perspectives are significant ($p>0.05$).
    \begin{table*}[!htbp]
        \centering
        \small
        \begin{tabular}{| P{1.5cm} | P{.5cm} P{.5cm} P{.5cm} P{.5cm} P{.5cm} P{.7cm} P{.7cm} || P{.5cm} P{.5cm} P{.5cm} P{.5cm} P{.5cm} P{.7cm} P{.7cm} | }
            \hline
            \multirow{2}{1cm}{\textbf{Model}} & \multicolumn{7}{ c ||}{Left} & \multicolumn{7}{ c |}{Right} \\ 
            & S & O & I & C & R & Avg. Len & Avg. Ext & S & O & I & C & R & Avg. Len & Avg. Ext \\
            \hline
            Ref.      & 2.9 & 3.1 & 2.8 & 2.9 & 1.7 & 19.4 & 0.65 & 2.9 & 3.4 & 2.9 & 2.9 & 2.0 & 20.0 & 0.69 \\
            \hline
            Mistral   & 2.4 & 3.1 & 2.5 & 2.6 & 2.0 & 47.1 & 0.83 & 2.4 & 3.2 & 2.5 & 2.5 & 1.8 & 50.0 & 0.84 \\
            Mixtral   & 3.2 & 3.4 & 3.1 & 3.2 & 2.7 & 45.7 & 0.71 & 2.8 & 3.4 & 2.8 & 2.8 & 2.4 & 46.8 & 0.73 \\
            GPT-4o    & \textbf{3.3} & \textbf{3.5} & \textbf{3.3} & \textbf{3.3} & \textbf{2.9} & 40.0 & 0.65 & \textbf{3.2} & \textbf{3.6} & \textbf{3.2} & \textbf{3.2} & \textbf{2.8} & 39.5 & 0.67 \\
            \hline
        \end{tabular}
        \caption{Average human judgments of each perspective dimension with average length and extraction coverage over all inputs. Scores are shown separately for left and right-perspective summaries. Best scores are \textbf{bolded}. Higher scores reflect better performance for all metrics. Length is measured in Llama-3 tokens. \textit{S}: Stance, \textit{O}: Object, \textit{I}: Intensity, \textit{C}: Consensus, \textit{R}: Reasoning.}
        \label{tab:human-judge-full}
    \end{table*}

\subsection{Human Evaluation Interface}
\label{app:annot-inter}

\begin{figure*}
        \centering
        \includegraphics[width=1.0\linewidth]{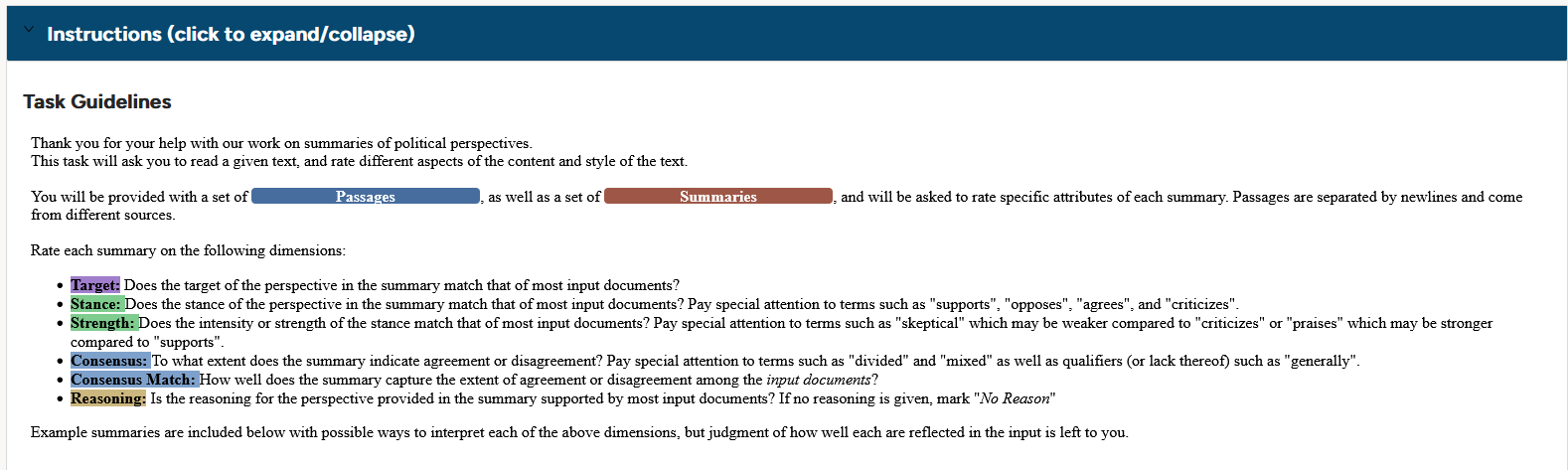}
        \caption{Human evaluation interface screenshot: Instructions.}
        \label{fig:annot-inter-1}
    \end{figure*}

\begin{figure*}
        \centering
        \includegraphics[width=1.0\linewidth]{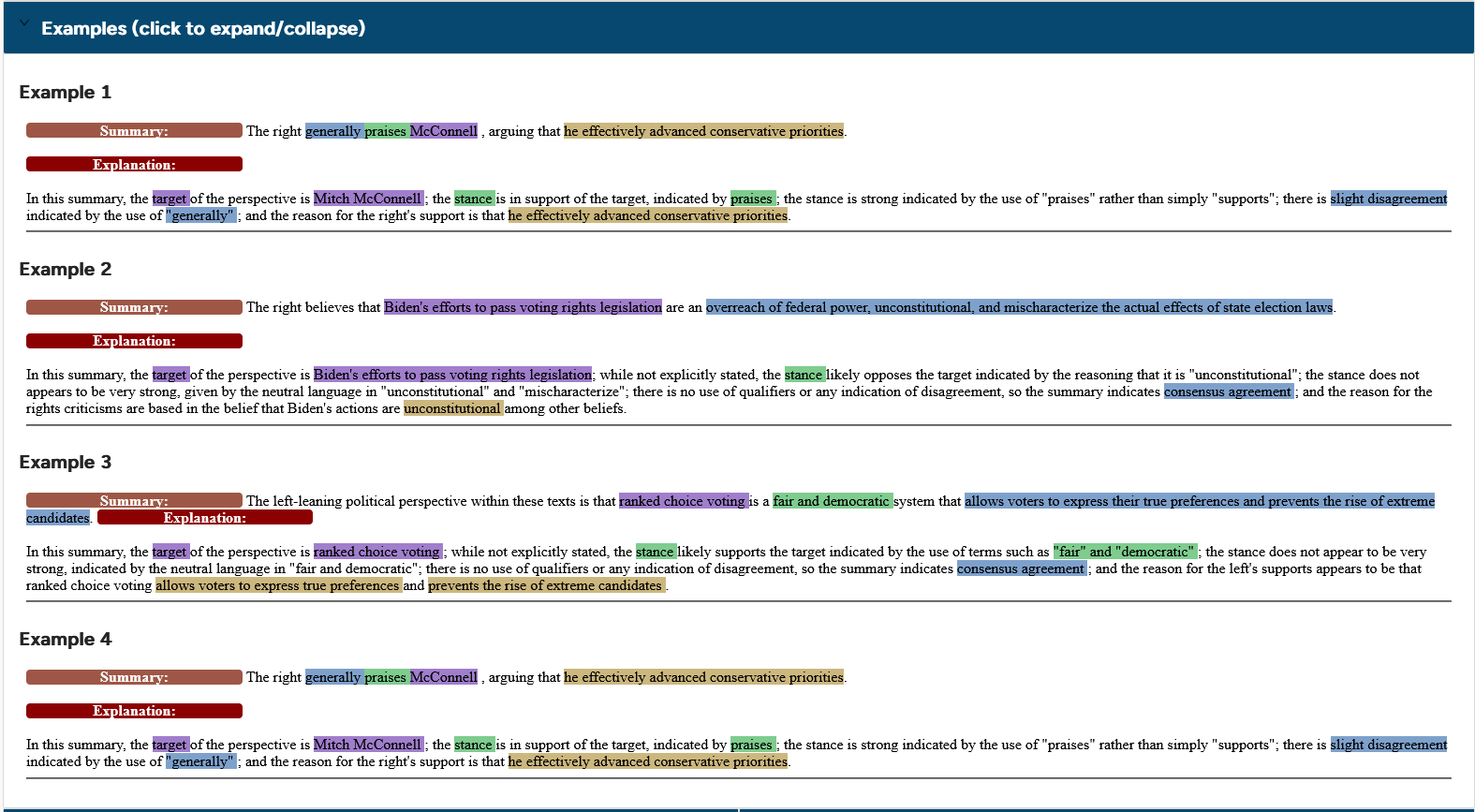}
        \caption{Human evaluation interface screenshot: Examples.}
        \label{fig:annot-inter-2}
    \end{figure*}

\begin{figure*}
        \centering
        \includegraphics[width=1.0\linewidth]{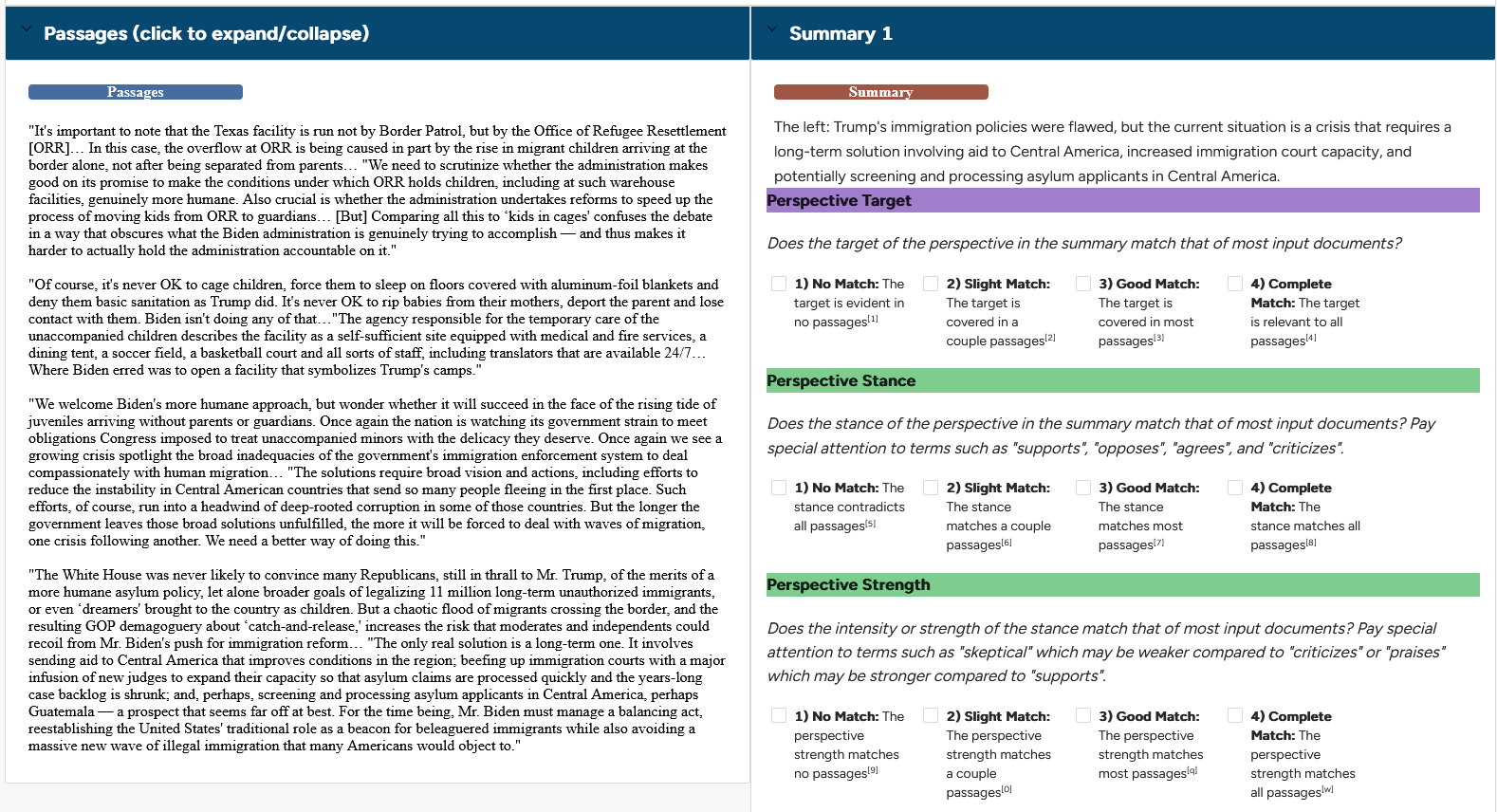}
        \caption{Human evaluation interface screenshot: Passages and Questions Part 1.}
        \label{fig:annot-inter-3}
    \end{figure*}

\begin{figure*}
        \centering
        \includegraphics[width=1.0\linewidth]{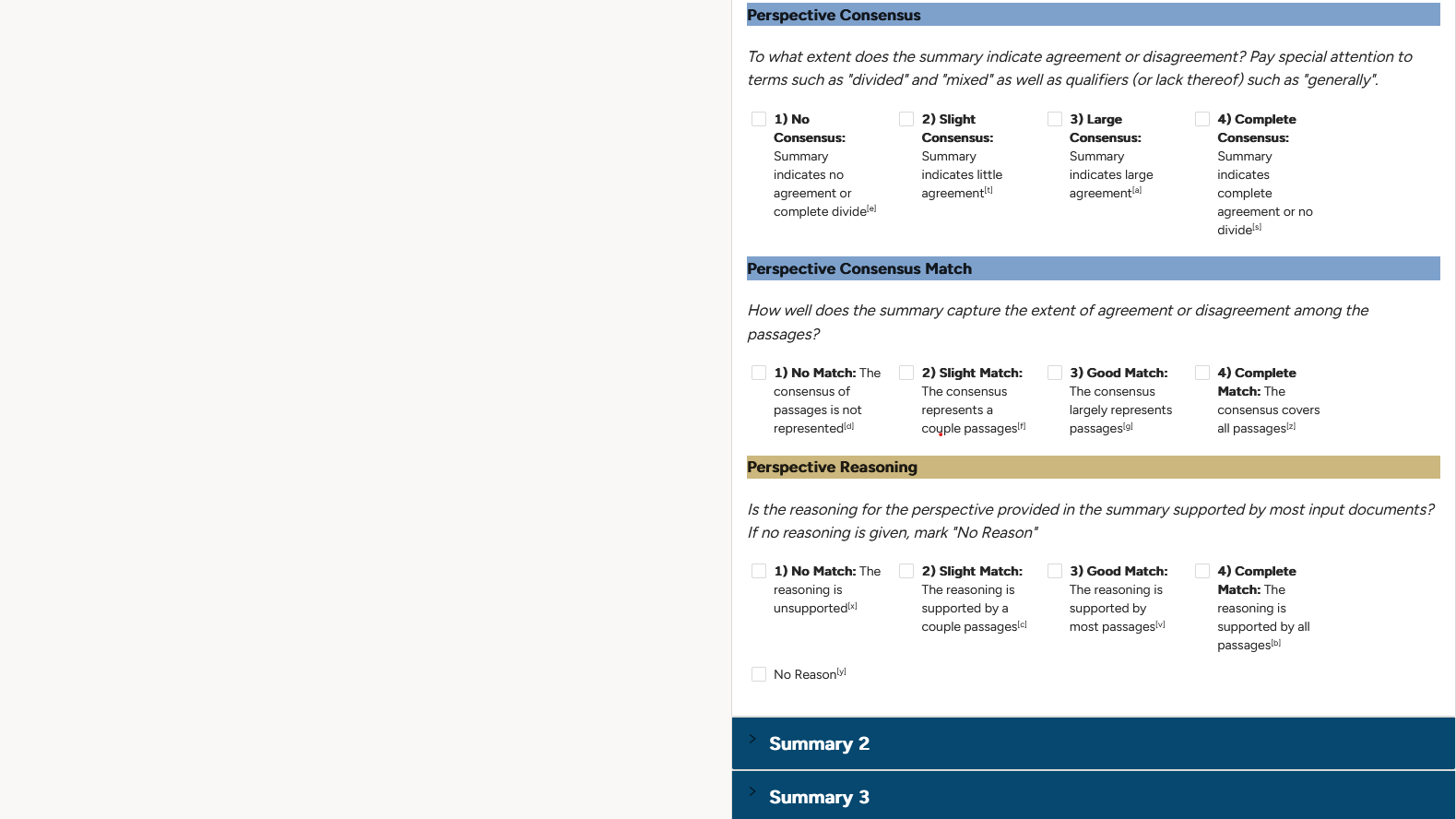}
        \caption{Human evaluation interface screenshot: Passages and Questions Part 2.}
        \label{fig:annot-inter-4}
    \end{figure*}

    Screenshots of the human evaluation interface using Label Studio \cite{label-studio} provided to annotators are shown in \autoref{fig:annot-inter-1}, \autoref{fig:annot-inter-2}, \autoref{fig:annot-inter-3}, and \autoref{fig:annot-inter-4}. Annotators are first provided with instructions describing each perspective dimension (\autoref{fig:annot-inter-1}). Annotators are then shown example summaries with accompanying explanations describing how each perspective dimension could be interpreted in the example summary (\autoref{fig:annot-inter-2}). We do not include input documents or possible judgments for the example summaries to avoid biasing annotators' judgments. In the left-hand column, annotators are given the input passages separated by new lines, and in the right-hand column, annotators are provided with a summary followed by questions and 4-point Likert scales for each perspective dimension (\autoref{fig:annot-inter-3} \& \autoref{fig:annot-inter-4}). Reference summaries, and summaries generated by Mistral, Mixtral, and GPT-4o are shown on the same page in a randomized order for each set of input passages. Annotators are able to collapse and un-collapse all sections.

\section{Coverage and Faithfulness Metric Correlations}
\label{app:summ-corr}

    \autoref{tab:cov-corrs} presents correlations between human judgments on Mistral, Mixtral, and GPT-4o compared to each coverage and faithfulness metric. Among coverage metrics, \textsc{Bleurt} and BERTScore tend to be best correlated among left-perspective summaries, while \textsc{Bleurt} and ROUGE metrics tend to be best correlated among right-perspective summaries. The low correlation coefficients overall and particularly for faithfulness metrics, however, suggest that further investigation is needed to adapt coverage and faithfulness metrics to this domain.  

    \begin{table*}[!htbp]
        \centering
        \scriptsize
        \begin{tabular}{| P{1cm} | P{.6cm} P{.6cm} P{.9cm} P{.8cm} | P{.9cm} P{.8cm} || P{.6cm} P{.6cm} P{.9cm} P{.8cm} | P{.9cm} P{.8cm} |}
            \hline
            \multirow{3}{1cm}{\centering {Judge \\ Dimension}} & \multicolumn{6}{c||}{Left} & \multicolumn{6}{c|}{Right} \\
            & \multicolumn{4}{c|}{Coverage} & \multicolumn{2}{c||}{Faithfulness} & \multicolumn{4}{c|}{Coverage} & \multicolumn{2}{c|}{Faithfulness} \\
            & R1   & RL   & \textsc{Bleurt} & BERT& \textsc{SummaC} & Align & R1   & RL    & \textsc{Bleurt} & BERT & \textsc{SummaC} & Align \\
            \hline
            Stance      & -0.03 & -0.06 & 0.22$^*$ & 0.14$^*$   & 0.06  & 0.19$^*$  & 0.10      & 0.12      & 0.24$^*$ & 0.00 & 0.02  & 0.15$^*$ \\
            Object      & 0.02  & -0.01 & 0.25$^*$ & 0.02       & 0.08  & 0.13      & 0.10      & 0.13      & 0.18$^*$ & 0.11 & 0.07  & -0.01 \\
            Intensity   & 0.05  & -0.02 & 0.27$^*$ & 0.21$^*$   & 0.05  & 0.20$^*$  & 0.09      & 0.13      & 0.26$^*$ & 0.09 & -0.01 & 0.12 \\
            Consensus   & 0.01  & -0.03 & 0.24$^*$ & 0.12       & 0.01  & 0.18$^*$  & 0.17$^*$  & 0.18$^*$  & 0.27$^*$ & 0.02 & -0.01 & 0.14$^*$ \\
            Reasoning   & -0.03 & -0.06 & 0.20$^*$ & 0.14$^*$   & -0.01 & 0.09      & 0.08      & 0.07      & 0.29$^*$ & 0.01 & -0.00 & 0.15$^*$ \\
            \hline
        \end{tabular}
        \caption{Kendall's Tau correlations between the 5 perspective dimensions measured with human judgments, and the automatic coverage and faithfulness metrics. * indicates significant correlation ($p\leq0.05$)}
        \label{tab:cov-corrs}
    \end{table*}

\section{Extraction Analyses by Perspective}

\begin{figure*}[!htbp]
        \centering
        \includegraphics[width=1.0\linewidth]{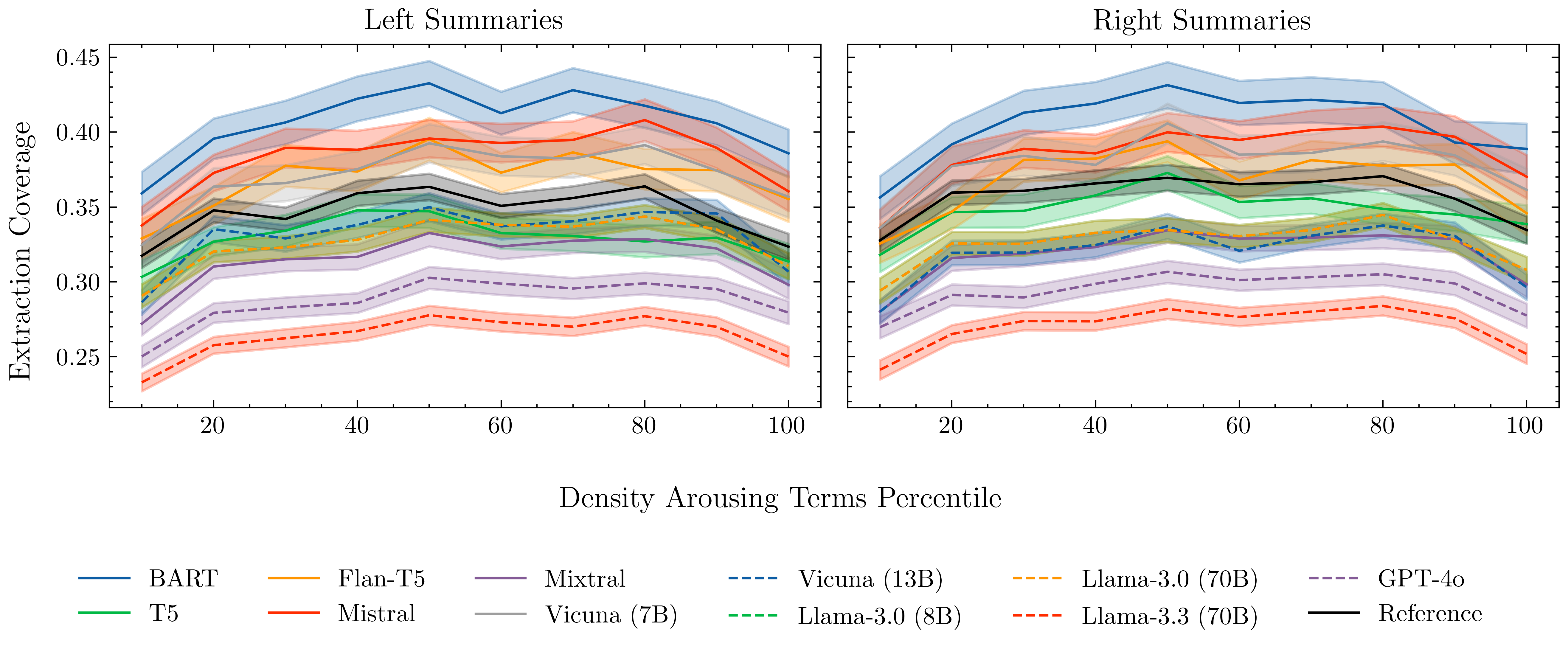}
        \caption{Average extraction coverage of source documents for each model by the number of highly arousing terms in the source document (percentiles are shown rather than raw values). }
        \label{fig:pos-extract-p}
    \end{figure*}

\begin{figure*}[!htbp]
        \centering
        \includegraphics[width=1.0\linewidth]{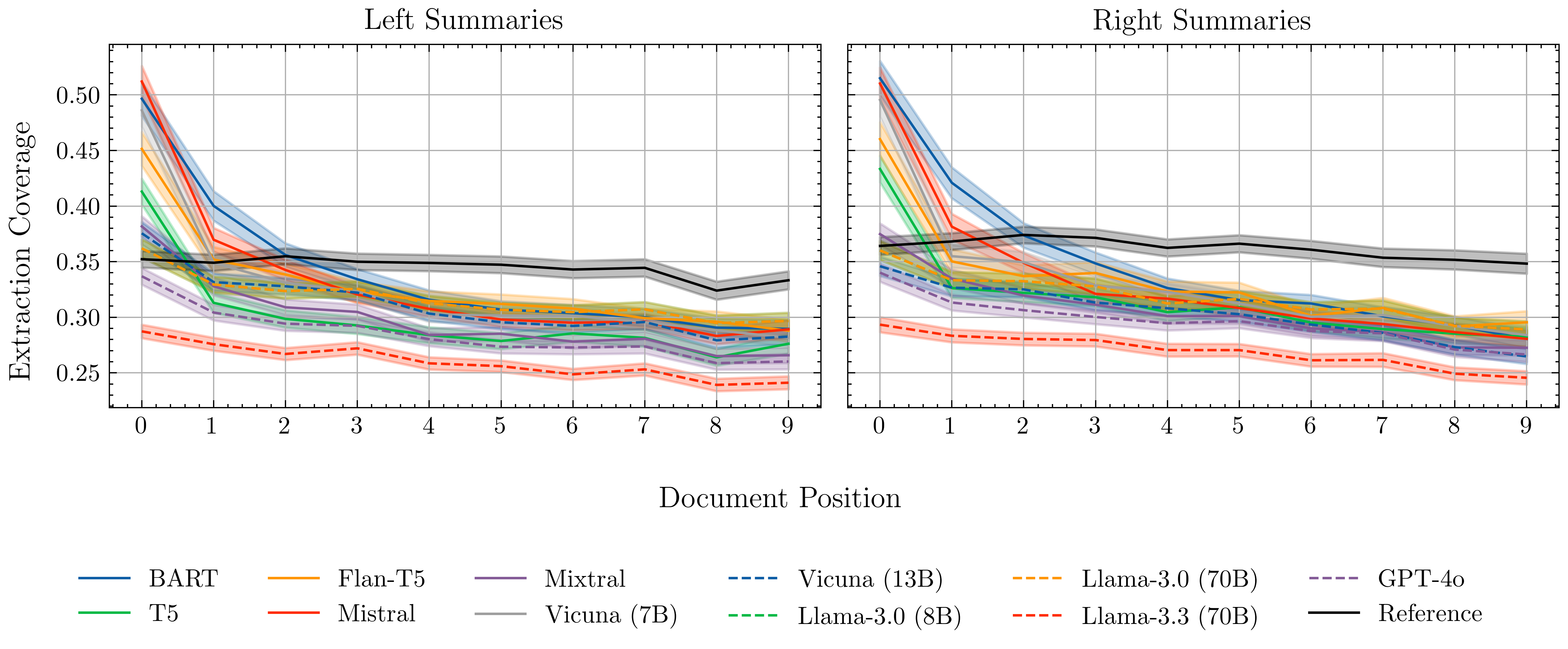}
        \caption{Average extraction coverage of source documents for each model by the document's position in the input. Lower values reflect earlier positions in the input.}
        \label{fig:len-extract-p}
    \end{figure*}

\begin{figure*}[!htbp]
        \centering
        \includegraphics[width=1.0\linewidth]{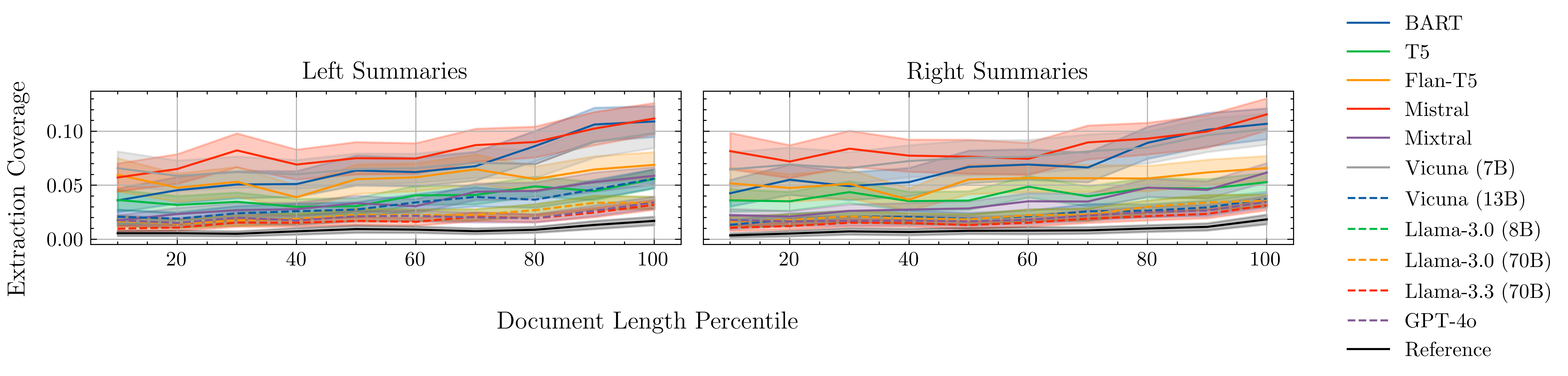}
        \caption{Average extraction coverage of source documents for each model by length of the input document (percentiles are shown rather than raw values). }
        \label{fig:narous-extract-p}
    \end{figure*}

    In \autoref{fig:pos-extract-p}, \autoref{fig:len-extract-p}, and \autoref{fig:narous-extract-p}, we present the position, length, and arousal bias analyses by perspective. All identified trends across the left and right perspectives are largely consistent with those reported in the results.

\section{Additional Summary Examples}
\label{app:add-summs}

    Additional summary examples are shown in \autoref{tab:add-summ-1}, \autoref{tab:add-summ-2}, and \autoref{tab:add-summ-3}.

    \begin{table*}[!htbp]
        \centering
        \small
        \begin{tabular}{| P{1.2cm} | P{6cm} | P{6cm} |}
            \hline
            \multicolumn{3}{|c|}{\textbf{Ron DeSantis}}  \\
            \hline
            Model & Left Summary & Right Summary \\
            \hline
            \textit{Reference} & The left argues that DeSantis’s campaign was doomed by miscues and strategic errors. & The right argues that Republican voters were just not ready to move on from Trump.	\\
            \Xhline{2pt}
            Mistral & The left: DeSantis' failure to \textit{{\color{blue} support aid for Ukraine and his signing of a six-week abortion ban in Florida alienated some moderates and made him look weak, ultimately hurting his chances of winning over voters open to a Trump alternative.}} & The right: \textit{{\color{violet}DeSantis's 2022 re-election campaign was successful}}, but his 2024 presidential campaign was unsuccessful due to his failure to win over {\color{blue} voters who were looking for a fresh start after the Trump-led defeats of 2018, 2020.} \\
            \hline
            Mixtral & The left perceives DeSantis as a weak candidate who failed to effectively differentiate himself from Trump, \textit{{\color{blue}ultimately losing support from both moderates and evangelicals}}, and struggling to manage his campaign and image, leading to his downfall in the 2024 GOP presidential primary. & The right believes that DeSantis, despite his initial potential, failed to effectively challenge Trump in the 2024 GOP presidential primary due to his inability \textit{{\color{red}to unite a broad coalition, lack of a clear message}}, and {\color{blue} Trump's political martyrdom following multiple indictments} \\
            \hline
            GPT-4o & The left views DeSantis' struggles as indicative of the broader Republican Party's inability to move beyond Trumpism, highlighting internal conflicts, ineffective strategies, and \textit{{\color{blue} the alienation of moderate voters}}. & The right views DeSantis' struggles as a reflection of the Republican base's enduring loyalty to Trump, despite his flaws, and the challenges of presenting a viable alternative within the party. \\
            \hline
        \end{tabular}
        \caption{Additional right-perspective example summaries on the most recent topic in the collected dataset (Ron DeSantis, January 2024). Distinct colors represent quotes or paraphrases from the same document, and italics denote information from passages reflecting the alternate perspective.}
        \label{tab:add-summ-1}
    \end{table*}

    \begin{table*}[!htbp]
        \centering
        \small
        \begin{tabular}{| P{1.2cm} | P{6cm} | P{6cm} |}
            \hline
            \multicolumn{3}{|c|}{\textbf{Race and Coronavirus}}  \\
            \hline
            Model & Left Summary & Right Summary \\
            \hline
            \textit{Reference} & The left argues that it is important to focus on racial disparities, because they stem from structural inequality organized along racial lines, and emphasizes the importance of societal solutions. & The right recommends focusing on the factors that underlie racial disparities, such as poverty and pre-existing health conditions, rather than race itself, and emphasizes the importance of individual responsibility. \\
            \Xhline{2pt}
            Mistral & The left: {\color{blue} \textit{The left argues that the outsized number of deaths in the black community is a serious issue that highlights the nation's failures in serving that community for decades.}} & The right argues that {\color{blue} the outsized number of deaths in the black community is a serious issue that points to the nation's failures in serving that community for decades, but it should not be the focus of the immediate response to the coronavirus.} \\
            \hline
            Mixtral & The left argues that systemic racial and socioeconomic disparities have led to {\color{violet} disproportionate exposure to air pollution and health impacts}, including COVID-19, {\color{violet} among black and Hispanic communities}, and that {\color{violet} immediate action such as expanding federal financial coverage for testing and treatment, prioritizing hospitals in at-risk neighborhoods}, and addressing underlying issues like  \textit{{\color{blue} economic and educational disparities}} is necessary. & The right perspective suggests that {\color{blue}while racial disparities in healthcare are a serious issue, the immediate response to the coronavirus should focus on getting medical resources to where they are needed most, regardless of race.} \\
            \hline
            GPT-4o & The left emphasizes the need to address systemic racial and \textit{{\color{blue} economic inequalities that exacerbate health disparities,}} advocating for immediate and \textit{{\color{blue}long-term measures to ensure equitable access to healthcare and resources for marginalized communities.}} & The right emphasizes {\color{red}individual responsibility and behavioral factors in health disparities}, {\color{blue} arguing that immediate pandemic responses should focus on actual risk factors rather than racial generalizations.} \\
            \hline
        \end{tabular}
        \caption{Additional example summaries for a randomly sampled article. Distinct colors represent quotes or paraphrases from the same document, and italics denote information from passages reflecting the alternate perspective.}
        \label{tab:add-summ-2}
    \end{table*}

        \begin{table*}[!htbp]
        \centering
        \small
        \begin{tabular}{| P{1.2cm} | P{6cm} | P{6cm} |}
            \hline
            \multicolumn{3}{|c|}{\textbf{Andrew McCabe}}  \\
            \hline
            Model & Left Summary & Right Summary \\
            \hline
            \textit{Reference} & The left pushes back on the narrative that exercising the 25th amendment would constitute a coup, and criticizes Republicans for hypocrisy regarding the ongoing investigations. & The right condemns McCabe and sees his statements as evidence that the investigation against Trump was politically motivated. \\
            \Xhline{2pt}
            Mistral & The left: {\color{blue} McCabe's revelations} suggest that Trump's firing of Comey was politically motivated and that {\color{blue}Nunes' objections to the FBI investigation may have been driven by bias.} & The right: \textit{{\color{blue}McCabe's revelations suggest that the ongoing FBI investigation of Trump and his campaign may have more to do with politics than the merits of the case, and that some politicians, including Nunes, were aware of the investigation and still worked with the White House to push a conspiracy theory.}} \\
            \hline
            Mixtral & The left perceives that top FBI officials, including McCabe, were concerned about {\color{violet} Trump's potential threat to national security} due to his alleged ties with Russia, and they believe that {\color{violet}certain Republican leaders, such as Nunes, McConnell, and Ryan, chose to overlook this concern and defend Trump instead.} & The right believes that the FBI, under the leadership of Andrew McCabe and James Comey, {\color{red}overstepped its authority by attempting to act as a check on President Trump}, \textit{{\color{blue}potentially influenced by political bias}}, and {\color{red}that the use of the 25th Amendment to remove Trump from office was an inappropriate and unrealistic measure.} \\
            \hline
            GPT-4o & The left views {\color{blue} McCabe's revelations} as evidence that top officials were justifiably concerned about {\color{blue}Trump's potential national security risks} and that the FBI's actions were a necessary check on presidential power. & The right views the actions of {\color{red} McCabe and other FBI officials as an overreach of their authority}, {\color{blue}driven by anti-Trump bias}, and believes that any attempt to remove Trump from office should follow constitutional processes rather than be decided by unelected officials. \\
            \hline
        \end{tabular}
        \caption{Additional example summaries for a randomly sampled article. Distinct colors represent quotes or paraphrases from the same document, and italics denote information from passages reflecting the alternate perspective.}
        \label{tab:add-summ-3}
    \end{table*}

\end{document}